# Public Computer Vision Datasets for Precision Livestock Farming A Systematic Survey

Anil Bhujel[a], Yibin Wang[b], Yuzhen Lu[b,*], Daniel Morris[b], Mukesh Dangol[a]
[a]*Ministry of Culture, Toursim and Civil Aviation, Kathmadu 44600, Nepal*
[b]*Department of Biosystems and Agricultural Engineering, Michigan State University, East Lansing, MI 48824, United States*

ARTICLEINFO

*Keywords:*
Animal agriculture
Computer vision
Data sharing
Precision livestock farming
Public datasets

ABSTRACT

Technology-driven precision livestock farming (PLF) empowers practitioners to monitor and analyze animal growth and health conditions for improved productivity and welfare. Computer vision (CV) is indispensable in PLF by using cameras and computer algorithms to supplement or supersede manual efforts for livestock data acquisition. Data availability is crucial for developing innovative monitoring and analysis systems through artificial intelligence-based techniques. However, data curation processes are tedious, time-consuming, and resource intensive. This study presents the first systematic survey of publicly available livestock CV datasets (https://github.com/Anil-Bhujel/Public-Computer-Vision-Dataset-A-Systematic-Survey). Among 58 public datasets identified and analyzed, encompassing different species of livestock, almost half of them are for cattle, followed by swine, poultry, and other animals. Individual animal detection and color imaging are the dominant application and imaging modality for livestock. The characteristics and baseline applications of the datasets are discussed, emphasizing the implications for animal welfare advocates. Challenges and opportunities are also discussed to inspire further efforts in developing livestock CV datasets. This study highlights that the limited quantity of high-quality annotated datasets collected from diverse environments, animals, and applications, the absence of contextual metadata, are a real bottleneck in PLF.

## 1. Introduction

Advances in artificial intelligence (AI) through deep learning (DL) (LeCun et al., 2015; Bengio et al., 2021) have reshaped the fields of computer vision (CV) and pattern recognition. The ImageNet Large Scale Visual Recognition Challenge (ILSVRC) (Russakovsky et al., 2015) provided one of the most widely used open-source, large-scale datasets, accelerating the CV community efforts to develop advanced network architecture towards human-level accuracy. The availability of large-scale public image datasets creates numerous opportunities for algorithmic achievements in various application domains. Today's DL-powered computer vision systems are well poised to supplant a majority of human labor while enabling 24/7 automatic monitoring task implementation in agricultural production. Through data-driven smart monitoring and intervention technologies, agricultural systems are transitioning to a new smart digital era with substantially increased production efficiency and reduced environmental impacts (Lu and Young, 2020).

Precision livestock farming (PLF) is an emerging and vibrant field that is nurtured heavily by computer vision and AI. In the past decade, considerable research has been conducted on using CV systems for high-throughput phenotyping, behavior monitoring, and welfare assessment of farm animals (Wurtz et al., 2019; Chen et al., 2021; Okinda et al., 2020; Li et al., 2022; Wang et al., 2023). Different imaging technologies are available for implementing CV systems in livestock production facilities, among which RGB (red-green-blue) or color imaging and RGB-D (depth) imaging are the most popular modalities. Particularly, consumer-grade RGB-D sensors for three-dimensional (3-D) CV systems have gained popularity for biometric body measurements (e.g., weight estimation) of livestock (Wang et al., 2023). In the past few years, there has been a paradigm shift in exploiting CV data acquired from livestock from traditional image analysis relying on feature engineering to data-driven DL approaches (Chen et al., 2021; Oliveira et al., 2021; Yousefi et al., 2022), leading to enhanced task performance towards practical application. There are certain prerequisites for DL to work effectively;

* Correspondence: *luyuzhen@msu.edu*

modern DL models, which contain millions or billions of learnable parameters, are inherently data-hungry to train, requiring large sets of annotated data. Creating large image datasets suitable for DL models, especially for training from scratch, is a notoriously laborious, tedious process, and sometimes is not viable in livestock production. There are well-known large-scale, publicly available datasets in the CV community, such as COCO (Lin et al., 2014) and ImageNet (Russakovsky et al., 2015). Yet, there is still a lack of such massive image datasets dedicated to PLF.

Numerous reviews dealing with CV and DL in livestock farming have been conducted in recent years. Nasirahmadi et al. (2017) were presumably among the first to summarize various CV systems applied for behavior detection of cattle and pigs. Okinda et al. (2020) analyzed various CV systems used for poultry welfare monitoring. Oliveira et al. (2021) highlighted DL algorithms and their development in various livestock applications such as image classification, object detection, segmentation, and phenotype detection. Similarly, the applications of DL in precision cattle farming were reviewed by Mahmud et al. (2021). Qiao et al. (2021) analyze different techniques and tools used for cattle detection, evaluation of cattle body condition scoring, and weight estimation. Arulmozhi et al. (2021) highlighted the use of video/images (2D and 3D) for behavioral monitoring of pigs and discussed the limitations facing in this field. Chen et al. (2021) analyzed the development process from CV to DL for behavior recognition of pigs and cattle. Different from previous reviews, Bahlo and Dahlhaus (2021) investigated the quality of open livestock datasets in Australia in terms of FAIRness (Findability, Accessibility, Interoperability, and Repeatability), but did not specify the percentage of identified image datasets, if any. Yousefi et al. (2022) presented a systematic review on the use of DL methods for livestock detection and localization using unmanned aerial vehicle (UAV) images, performance evaluation metrics, and databases. Dohmen et al. (2022) examined studies on the body weight measurement of livestock using CV and machine learning (ML), challenges and possible solutions as well as evaluation parameters for validating ML models. Recently, Wang et al. (2023) surveyed the applications of 3D CV and image pre-processing techniques, with an emphasis on cattle growth management.

Despite these previous reviews, they focused primarily on computer vision and ML/DL applications in PLF. Although some research reviewed image datasets for specific livestock species (Bahlo and Dahlhaus, 2021; Han et al., 2023; Hossain et al., 2022), they are short of delivering in-depth coverage, taxonomy, and evaluation of public datasets broadly on different livestock species. The PLF research community could have benefited greatly from open-source image datasets. Still, there has been no comprehensive study of public datasets in terms of applications, data volumes, image modalities, and imaging environments for important livestock species. This survey was therefore to fill the gap and systematically unearth the publicly available image/video datasets for a wide range of CV and ML/DL tasks in PLF. The main contribution of this survey was to identify, classify, and critically discuss characteristics of publicly available livestock datasets and future research needs in data curation and sharing.

The subsequent sections of this paper are structured as follows. **Section 2** describes the methodologies used to collect relevant studies and identify public livestock datasets therein. The main characteristics of the identified datasets are presented in **Section 3**, on camera sensors, imaging environments, data annotation and volumes, and applications. **Section 4** identifies limitations in previous work and discusses challenges and opportunities in creating open-source livestock datasets, followed by a concluding summary.

## 2. Methodology

To systematically survey public datasets in PLF, we followed the standard guidelines in the PRISMA 2020 statement (Page et al., 2021). With the objective of surveying publicly available livestock datasets, a set of search inclusion criteria and search keywords was identified to have comprehensive literature coverage while filtering out irrelevant articles. The following criteria were used to identify whether a specific paper should be included for detailed reporting regarding dataset relevance and availability.

- The data should be used for a computer vision application.
- The data should be publicly available without requiring the author's consent.
- The data should be used in any livestock phenotype study.

Additionally, it is found that open-source CV datasets came into existence mostly after the application of DL in mid-2010. Hence, only research papers published from 2015 until the time of writing are considered in this survey. Then, the appropriate keywords and digital journal archives were fixed to search for relevant research papers. Figure 1 shows the procedure followed while searching and filtering related articles for review.

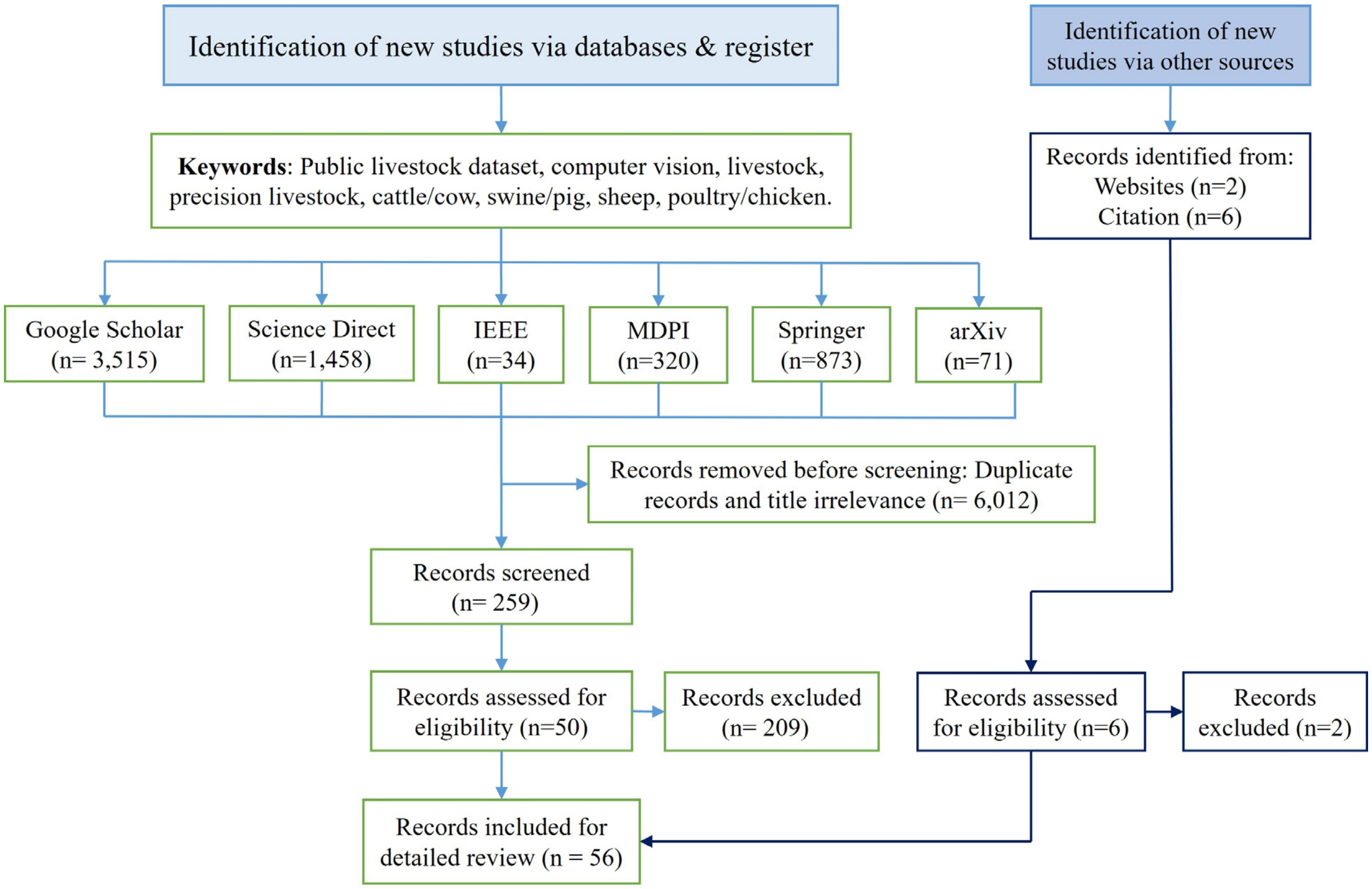

**Figure 1**. Preferred reporting items for systematic review and meta-analyses (Page et al., 2020).

Various keywords such as “livestock dataset”, “public livestock dataset”, “image dataset”, “livestock monitoring”, “cow/cattle monitoring”, “pig/swine monitoring”, “poultry/chicken monitoring”, “sheep monitoring”, etc., and their combinations were used in mainstreams archives, applying advanced search options. Then, the searched results were skimmed according to their title and relevance and narrowed down the results combining apposite keywords or phrases. In this way, 259 articles were found and further scrutinized and filtered out corresponding to research questions and refined them to 50. Moreover, 8 articles were obtained through other sources such as secondary citations and websites, which underwent a similar filtering process and were reduced to 6. Thus, a total of 56 research articles were finalized and rationally reviewed. The selected scholarly articles are further grouped as per the publication year to identify the trend of livestock dataset publication over time, as shown in Figure 2.

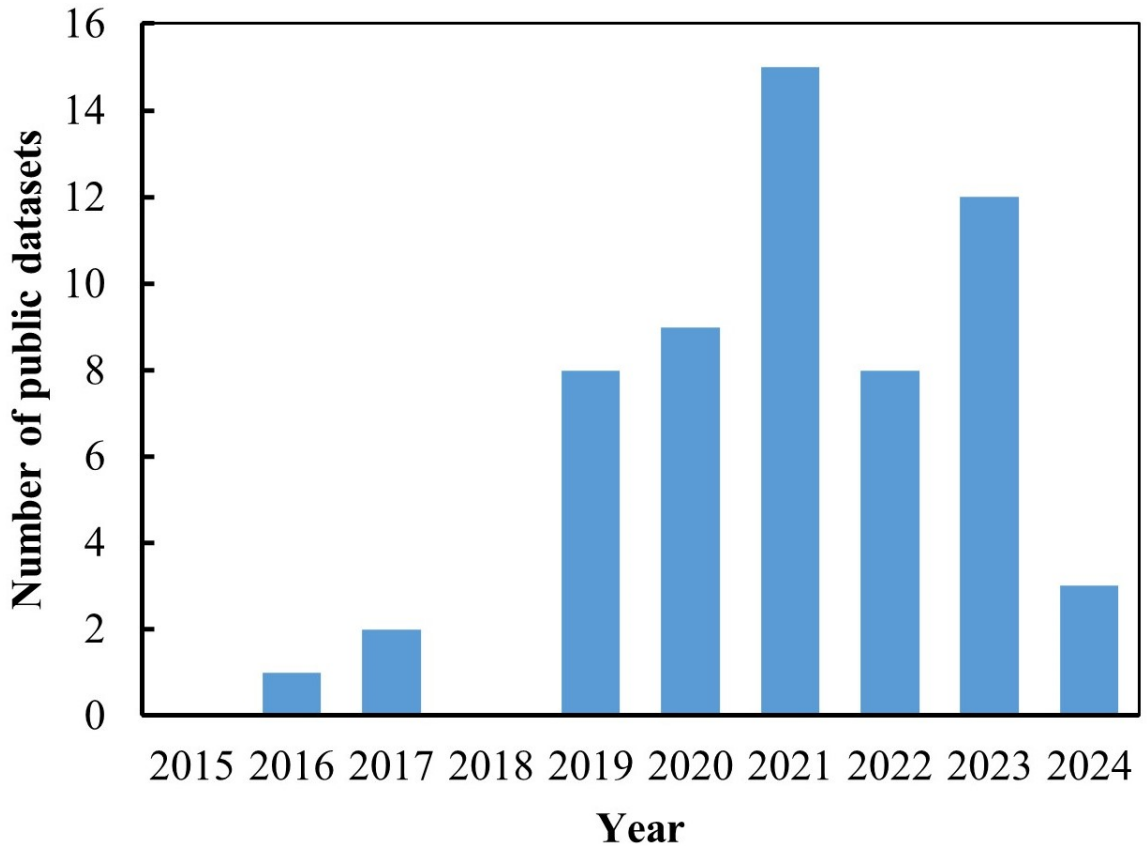

**Figure 2**. Number of public computer vision livestock datasets over the past ten years.

## 3. Datasets

There is unanimity among researchers on the scarcity of publicly available, high-quality CV datasets of livestock (Bahlo and Dahlhaus, 2021; Han et al., 2019; Ocholla et al., 2024; Ong et al., 2023; Psota et al., 2019; Vayssade et al., 2023). While the volume of livestock data in public sources continues to grow with additions from individuals and research institutions, it remains far below the scale of generic CV datasets like ImageNet and MS COCO. Presented below is a review of the publicly available livestock CV datasets published in the last decade, which are categorized according to animal species. A total of 58 datasets sourced from 56 published articles were examined and presented individually in this study. These datasets are freely available on the internet and the authors already provided their consent for reusing them. Many of these datasets are in readily usable conditions, whereas a few datasets are in raw or semi-processed form, requiring further processing before applying them for modeling. All dataset information can be accessed via our GitHub website[i].

### 3.1. Cattle Datasets

The predominant public livestock datasets focused on cattle. Table 1 shows 27 relevant datasets available with annotations. This section sheds light on the purpose and application of each dataset along with the associated imaging sensor and environment, and the size of the dataset.

#### *3.1.1. FriesianCattle2015*

The FriesianCattle2015 is the first 3D dataset captured and made public by Andrew et al. (2016). It was used for detecting individual Holstein Friesian cattle through their distinctive dorsal color patterns and markings. The cow objects were extracted from the background using empirically determined maximum and minimum (3.4 m and 2 m) distance thresholds. The preprocessing yielded 764 RGB-D image pairs of 92 individual cows and then annotated at the image level. The authors captured the dataset in a farm environment using a consumer-grade RGB-D camera mounted 4 m above the ground when the cow exited the milking station. The camera was configured at a depth resolution of 512×424 pixels represented by 16 bits and RGB video frame resolution at 1920×1080, recorded at 30 frames per second (fps). The authors divided the dataset into training and testing subsets, containing 83 images from 10 individual cows and 294 images from 40 individual cows, respectively.

#### *3.1.2. FriesianCattle2017*

Andrew et al. (2017) implemented a regional convolutional neural network (R-CNN) model to localize and detect individual cattle. The dataset was prepared by manually annotating cow objects with bounding boxes and assigning a unique ID to each cow. They created two datasets: 1) the FriesianCattle2017 was captured from a top view RGB camera located at the ground, and 2) the AerialCattle2017 was shot from an aerial view using a drone (Andrew et al., 2017). The FriesianCattle2017 consists of 940 images from 89 individuals in an indoor environment captured by an RGB-D camera mounted over the passage between holding pens and the milking station. Indeed, the camera and setup were the same as the FriesianCattle2015 dataset (Andrew et al., 2016).

#### *3.1.3. AerialCattle2017*

The AerialCattle2017 dataset was gathered from an outdoor field environment using a drone equipped with an RGB camera (Andrew et al., 2017). Like the FriesianCattle2017 dataset, it was also used for individual cow localization and detection by applying an object detection model. The ground-truth labeling of cow objects was done manually using bounding boxes for the first frame of a video. Then, the Kernelized Correlation Filters (KCF) tracking algorithm (Henriques et al., 2015) was applied to the subsequent frames of the video to identify and crop the cow's region of interest (ROI) and finally labeled the cropped ROI images into respective cow classes. The AerialCattle2017 captured 34 herd video clips with an average of approximately 20 seconds duration. The acquired videos were pre-processed, resulting in 46,340 RGB frames for public access.

#### *3.1.4. Aerial-livestock-dataset*

The Aerial-Livestock-Dataset was collected over the grassland using a quadcopter with the aim of livestock detection (Han et al., 2019). This dataset consists of 89 aerially filmed images with a high resolution (3000 × 4000 pixels), and each livestock instance is labeled "livestock" with a bounding box, creating 4996 instances. It has three subcategories according to the livestock colors and background complexity, as shown in Figure 3. Category I contains 29 images of black-colored livestock with a simple background. Category II also consists of 17 frames of black-colored cows with geomorphologically noisy backgrounds. Category III consists of the most challenging 43 images having densely populated, multi-colored (black, grey, white, etc.) livestock with other confounding factors such as snow, buildings, land structure, etc.

**Table 1**. Public cattle datasets.

| Dataset | Modality | Imaging environment | Dataset size | Annotation | Applications | URL |
|---|---|---|---|---|---|---|
| FriesianCattle2015 (Andrew et al., 2016) | RGB-D | Farm milking station | 83 training and 294 testing images | Image-level | Individual cattle detection | https://data.bris.ac.uk/data/dataset/wurzq71kfm561ljahbwjhx9n3 |
| FriesianCattle2017 (Andrew et al., 2017) | RGB | Farm | 940 RGB images | Bounding box | Individual cattle detection | https://data.bris.ac.uk/data/dataset/2yizcfbkuv4352pzc32n54371r |
| AerialCattle2017 (Andrew *et al*. 2017) | RGB | Outdoor field | 46,430 jpg frames | Bounding box | Individual cattle detection | https://data.bris.ac.uk/data/dataset/3owflku95bxsx24643cybxu3qh |
| Aerial-livestock-dataset (Han et al., 2019) | RGB | Outdoor field | 89 images with 4996 instances | Bounding box | Cattle detection | https://github.com/ hanl2010/Aerial-livestock-dataset/releases |
| BeefCattleMuzzle (G. Li et al., 2022) | RGB | Open environment | 4,923 images | Image-level | Cattle muzzle detection | http://doi.10.5281/zenodo.6324360. |
| OpenCows2020 (Andrew et al., 2021) | RGB | Indoor and Outdoor field | 3,703 images with 6,917 instances | Bounding box | Individual cattle detection | https://data.bris.ac.uk/data/dataset/10m32xl88x2b61zlkkgz3fml17 |
| Cows2021 (Gao et al., 2021) | RGB | Outdoor field | 10,402 images | Bounding boxes with box orientation angle | Individual cattle detection | https://data.bris.ac.uk/data/dataset/4vnrca7qw1642qlwxjadp87h7 |
| Holstein CowRecognition (Bhole et al., 2019) | Thermal and RGB | Research farm | 1,237 pairs of thermal and RGB images | Image-level | Individual cattle detection | https://dataverse.nl/dataset.xhtml?persistentId=doi:10.34894/O1ZBSA |
| HolsteinCattle (Bhole et al., 2022) | Thermal and RGB | Research farm | 3,694 pairs of RGB and Thermal | Image-level | Individual cattle detection | https://dataverse.nl/dataset.xhtml?persistentId=doi:10.34894/7M108F |
| RecBov51c Dataset (Weber et al., 2020) | RGB | Farm environment | 27,849 images | Image-level | Cattle breed detection | https://data.mendeley.com/datasets/8ysxtyf8p2/1 |
| CowBehavior (Koskela et al., 2022) | RGB | Milking Station | > 1.7 million images | Image-level | Cow behavior detection | https://zenodo.org/record/3981400#.ZDVEXHZBxPY |
| NWAFU_CattleDataset (Li et al., 2019) | RGB | Outdoor field | 2,134 images form 63 cattle | Landmark (16 body key points) | Cattle pose detection | https://drive.google.com/open?id=1bLQFHd9rqllmEaYvbcAqEBJtlyfuKPdP |
| CowDB (Ruchay et al., 2020) | RGB-D | Commercial farm | 154 Hereford cattle | N/A | Cattle body measurement | https://github.com/ruchaya/CowDB |
| CowDatabase (Ruchay et al., 2020) | RGB-D | Commercial farm | 103 Hereford cattle | N/A | Body measurement | https://github.com/ruchaya/CowDatabase |
| CowDatabase2 (Ruchay et al., 2022b) | RGB-D | Commercial farm | 119 Black Angus cow | N/A | Body measurement | https://github.com/ruchaya/CowDatabase2 |
| 300-Cattle-Source (Shojaeipour et al., 2021) | RGB | Head scooping system | 2,632 images from 300 cattle | Image-level | Cattle muzzle detection | https://cloud.une.edu.au/index.php/s/eMwaHAPK08dCDru |
| CattleVideo (Qiao et al., 2021a) | RGB | Commercial farm | 14,400 images training and 3,750 for testing | Image-level | Tracking by detection | https://drive.google.com/file/d/13LZmzb5XcqzUVQo3EiTIf4pur4lFHmZD/view?usp=sharing |
| MultiviewC (Ma et al., 2021) | RGB-D | Commercial farm | 3,920 images | 3D bounding box | 3D cattle detection | https://github.com/Jiahao-Ma/MultiviewC |
| Cattle-counting (Soares et al., 2021) | RGB | Large open fields | 5,058 images | Bounding boxes | Counting by detection | https://vhasoares.github.io/downloads.html |
| Cattle_Dataset (Z. Li et al., 2022) | RGB | Open environment | 10,239 Cow face images | Image-level | Cattle face detection | https://github.com/Ol iver6999/cattle_dataset/tree/main |
| Aerial Pasture (Shao et al., 2020) | RGB | Outdoor pasture area | 670 images with 1,948 annotations | Bounding box | Counting by detection | http://bird.nae-lab.org/cattle/ |
| Cattle_Video_Behaviors (CVB) (Zia et al., 2023) | RGB | Outdoor field | 502 fifteen-Sec video clips (450 frames each) | Bounding box | Behaviour detection | https://data.csiro.au/collection/csiro%3A58916v1 |
| LShapeAnalyser-cow (Zhang et al., 2023) | RGB-D | Indoor milking station | 203 Cows | N/A | Body measurement | https://gitee.com/kznd/lshape-analyser/tree/master/Dataset/dairy |
| PCD Dataset (Hou et al., 2023) | RGB-D | Breeding farm | 2,106 PCDs after augmentation from 100 Beef Cattle | Landmark (Back region, chest region, hip region, and blank region) | Body weight estimation | https://github.com/colorful-days/0724/tree/master/data |
| CattleEyeView (Ong et al., 2023) | RGB | Outdoor loading ramp in the farm | 14 top-down video sequences (30,703 frames) | Bounding box, semantic segmentation | Multi-tasking (detection, counting, pose estimation, tracking and semantic segmentation) | https://github.com/AnimalEyeQ/CattleEyeView?tab=readme-ov-file |
| Dairy Cow (Gong et al., 2022) | Infrared (IR) | Farm environment | 1800 sampled images from 10 cows | Bounding box, Landmark | Posture (Standing, Walking, and Lying) estimation | https://www.kaggle.com/twisdu/dairycow |
| Cows Frontal Face (Ahmed et al., 2024) | RGB | Different farms | 2893 images from 459 cows | bounding box | Cattle face detection | https://zenodo.org/records/10535934 |

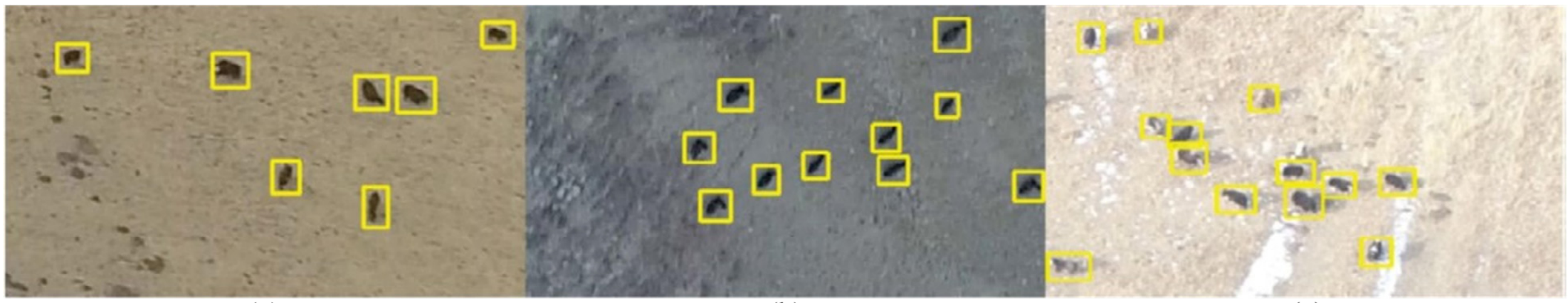

(a) (b) (c)

**Figure 3**. Sample images and annotations of three different categories (Han *et al*., 2019): (a) category I: simple; (b) category II: complex background; (c) category III: the most challenging.

### *3.1.5. BeefCattleMuzzle*

The BeefCattleMuzzle dataset was prepared by imaging beef cattle outside the housing pen taken from various distances using an RGB mirrorless digital camera (X-T4, FUJIFILM, Tokyo, Japan). The muzzle pattern, one of the biometric signatures of animals, was used in this dataset to identify individual beef cattle (Li et al., 2022). The data was collected when the animals were in natural behaviour conditions without human intervention. The acquired images were subjected to various preprocessing steps (focusing, cropping, and filtering). After the preprocessing, 4,923 annotated (image-level) muzzle images from 268 individual beef cattle remained in the dataset, with muzzle images of 300×300 pixels resolutions and segregated cattle-wise.

### *3.1.6. OpenCows2020*

This dataset contains the top-view RGB images of Holstein-Friesian cow images taken from both indoor and outdoor environments. The dataset was labeled manually using a bounding box covering the animal trunk (excluding the head, neck, and tail). It was used for the detection and biometric identification of Holstein-Friesian cows. Andrew et al. (2021) prepared this dataset by integrating the image data used in their previous works (Andrew et al., 2019, 2017, 2016), creating a total of 3,703 images with 6,917 cattle instances. The dataset contains diverse annotation box sizes, with only one or two animals in the majority of frames.

### *3.1.7. Cows2021*

The Cows2021 dataset was also applied to localize and detect Holstein-Friesian cows at individual levels. All cow instances in a video sequence were uniquely assigned one of the IDs among 186 cows (Gao et al., 2021). The authors utilized the dataset to validate a self-supervised network using contrastive learning and clustering. Volume-wise, the dataset contains 10,402 RGB images extracted from 301 video clips of 186 Holstein-Friesian cattle. The videos were shot using an RGB-D camera (RealSense D435, Intel Corporation, California, USA) in the field environment pointing downwards from 4 m height in the cattle walkway between pens and the milking parlor. The cattle objects were annotated, including only the body part and excluding the head, neck, tail, and leg portions, resulting in a total of 13,784 instances. Additional information ($\theta$) was added to represent the bounding box orientation corresponding to the centroid of the bounding box ($c_x$, $c_y$, $w$, $h$, $\theta$).

### *3.1.8. HolsteinCattleRecognition*

The purpose of this dataset (Bhole et al., 2019) was to recognize a single Holstein cow using thermal and RGB imaging. This dataset presents thermal and RGB image pairs of 136 milking cows, using a thermal camera (FLIR, Teledyne FLIR LLC, Wilsonville, USA), which was mounted 5 m apart from the animal at a milking station in the farm environment, capturing a side view of the animal. The minimum and maximum number of image pairs per animal are 6 and 14, respectively, for 136 Holstein cows, yielding 1,237 image pairs. The image pairs were annotated at the image level, assigning ID according to the collar tag attached to the animal.

### *3.1.9. HolsteinCattle*

Like the HolsteinCattleRecognition dataset (Bhole et al., 2019), the HolsteinCattle dataset (Bhole et al., 2022) was also created for the detection and monitoring of individual Holstein animals from a side view. This dataset used image-level annotation for categorical classification so that all the images belonging to individual cattle were stored separately. It contains 3694 pairs of RGB and thermal images from 383 individual Holstein cattle. The individual cow images were captured during every afternoon milking time at the milking parlour for 9 days. A thermal camera, mounted at a height of 1.5 m from the ground and 5 m away from the cattle, was used to capture the side view of the animal.

### 3.1.10. RecBov51c Dataset

This dataset was used for the recognition of Pantaneira breed-type cattle at the individual level (Weber et al., 2020). The images were taken from 51 Pantaneira cattle and annotated at the image level for identification. Both males and females of different age groups were captured using four RGB cameras, each located in various positions in the walkway of cattle on a commercial farm. Thus, the four cameras captured the different parts of the cattle body (back, profile, side, and face). In this way, a total of 27,849 frames were extracted from 212 video clips. Owing to the varying exposure time of each animal while taking the video, the dataset contains a different number of images per animal. This dataset has no full body images of cattle due to unavoidable obstruction on the head or leg areas. Images from each animal are stored separately, forming 51 classes for image-level classification.

### 3.1.11. CowBehavior

It is probably the largest cow behaviour dataset, containing more than 1.7 million images collected over two months (Koskela et al., 2022). This dataset was used to monitor the cow's behavior, identifying their various activities. A color camera, installed in front of an automatic milking station at 3.3m height, was used to capture the videos. Moreover, the camera position and inclination were changed to provide diversity in the dataset. The dataset contains 280 labelled videos (more than 1.7 million frames). On average, the length of each video was four minutes and nine seconds. Each video frame was annotated at the image level, assigning the corresponding class labels as given in Table 2.

**Table 2**. The classes used in labeling in their order of priority decreasing from top to bottom (Koskela et al., 2022)

| Code | Name | Brief description |
|---|---|---|
| -1 | undefined | Undefined situation or doubt about the correct class due to too much noise |
| 0 | human | There are human(s) in the image. |
| 1 | interaction frontal | Head on head or neck |
| 2 | interaction lateral | Head or body touching the other cow's head or body |
| 3 | interaction vertical | Head or other body part over body or head |
| 4 | crowded | More than five cows in the interest area |
| 5 | drinking | A cow is drinking water. |
| 6 | curiosity | Putting head inside the AMS housing. |
| 7 | queue | At least one cow is waiting outside the milking station. |
| 8 | low visibility | Night or twilight |
| 9 | normal | Nothing, laying down, peeing, etc. |

### 3.1.12. NWAFU-CattleDataset

The dataset is composed of 2,134 images of 63 cattle (33 dairy cattle and 30 beef cattle) annotated manually into 16 key points (Li et al., 2019). It was applied for detecting the poses of an animal in relation to animal behavior. The body landmarks labeled for pose estimation are 1-Head top, 2-Neck, 3-Spine, 4-Right front thigh root, 5-Right front knee, 6-Right front hoof, 7-Left front thigh root, 8-Left front knee, 9-Left front hoof, 10-Coccyx, 11-Right hind thigh root, 12-Right hind knee, 13-Left hind hoof, 14-Left hind thigh root, 15-Left hind knee, and 16-Left hind hoof. The videos were taken with a smartphone (at a resolution of 1920 × 1080 pixels and a rate of 30 fps) under natural conditions in a farmhouse.

### 3.1.13. CowDB

This dataset (Ruchay et al., 2020) was applied to reconstruct the 3-D body shape of a cow by using RGB and Depth information (two side views and a top view). The reconstructed shape was validated with the manually measured animal body shape. The RGB-D images were captured from the left, right, and top view of the 103 individual cattle, along with the manually measured nine body parameters (as shown in Figure 4). Two RGB depth cameras were installed on the right and left sides of the animal passage at approximately 2 m from the cattle and the top view camera at a height of 3 m from the ground.

### 3.1.14. CowDatabase

The CowDatabase dataset (Ruchay et al., 2020) consists of 154 Hereford cattle captured in the farm environment using an RGB-D camera with the aim of body shape estimation. It contains the manually measured ten body parameters of a cow (WT: Weight, WH: Wither Height, HH: Hip Height, CD: Chest Depth, HG: Heart

Girth, IW: Ilium Width, HJW: Hip Joint Width, OBL: Oblique Body Length, HL: Hip Length, CW: Chest Width), as shown in Figure 4, corresponding to each RGB-D image. The image resolution of depth and RGB images are 512 × 424 pixels and 1920 × 1080 pixels, respectively.

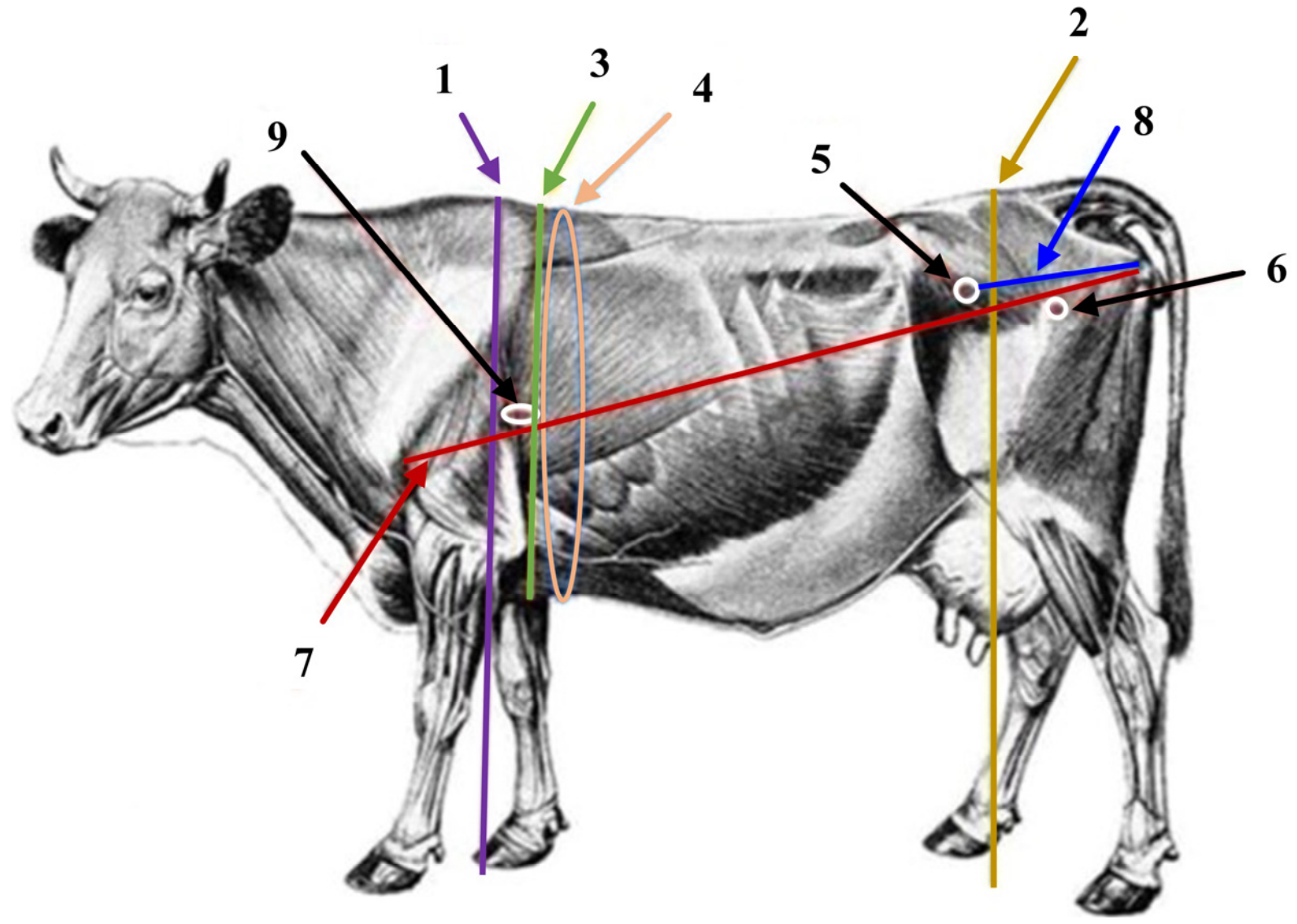

**Figure 4**. The nine manually measured body dimensions of cow: (1) Withers Height, (2) Hip Height, (3) Chest Depth, (4) Heart Girth, (5) Ilium Width, (6) Hip Joint Width, (7) Oblique Body Length, (8) Hip Length, and (9) Chest Width. Figure is adapted from Ruchay et al. (2020).

### *3.1.15. CowDatabase2*

Ruchay et al. (2022c) studied various production parameters, such as cow growth, productivity, and carcass quality, using genome-wide association studies and single nucleotide polymorphisms for the Aberdeen Angus cattle breed. They collected RGB and depth images from 96 individual cattle, including their physical tag number (RFID), body weight, withers height, hip height, chest width, chest height, genotyping data, and various slaughter characteristics (live weight, meat yield, stress losses, 19 characteristic parameters of the front quarters, 15 parameters of the rear quarters, and 22 of offal). Two Microsoft Kinect v2 depth cameras were installed around 2.0 m from the animal, and one from the top to collect three side point cloud data of a cow moving in a pass-way near the feeder. The image captured from multiple cameras was synchronized through a laptop, at 30 fps with an RGB and depth resolution of 1,920 × 1,080 and 512 × 424 pixels, respectively.

### *3.1.16. 300-Cattle-Source*

The dataset consists of 2,632 RGB images, after preprocessing, from 300 Bos Taurus beef cattle with mixed breeds (Shojaeipour et al., 2021). The author applied this dataset to train and validate a few-shot deep transfer learning model for individual cattle detection from their muzzle pattern. The images were captured under natural lighting conditions on the outdoor farm, with the camera placed 1 m above the ground when the cattle head remained in a head scoop during the eye treatment. The camera was configured in burst mode and clicked ten pictures (6 shots per second).

### *3.1.17. CattleVideo*

This dataset used 363 rear-view videos from 50 cattle shot at three different periods with a monthly interval (Qiao et al., 2021a). The video sequences were fed to the Inception-v3 model to extract features applied to the attention-based long short-term memory model for animal detection. A stereovision camera (ZED, StereoLabs, San Francisco, CA, USA), configured at a frame resolution of 1920 × 1080 pixels and 30 fps, was installed at the commercial farm to capture a video clip of 40 frames. The animal identity was manually labeled from video clips according to the ear tag. Since the videos were recorded in the natural farm conditions on the walkway to the feedlot at different dates and times, the dataset contains diverse animal postures, cluttered backgrounds, and varying lighting levels, possessing challenges to animal identification tasks.

### 3.1.18. MultiviewC dataset

Ma et al. (2021) prepared a synthetic multiview 3D dataset that allows realistic and diverse test cases for varying object sizes. The dataset was generated using Unreal Engine (v4.26) after fixing the scene on a cattle farm with sunny lighting conditions, covering 39 m × 39 m square of ground, quantizing into a 3,900 × 3,900 grid. The scene was captured using seven cameras, fixing 4 of them at four corners of the area and the other 3 cameras on the top of the drinking water trough. The camera was set up to capture an image of 720 × 1280 pixels every two seconds. Thus, 560 images were recorded from each camera, resulting in 3,920 in total, targeting 15 cows that demonstrated five random actions (sleeping, grazing, running, walking, and lazing). This dataset contains the diverse sizes of the object in all 3Ds when the animal is in different actions.

### 3.1.19. Cattle-counting

This dataset consists of 5,058 aerial images captured from large pasture areas from six farm areas with georeferenced information (Soares et al., 2021). The animals in the images were located and labeled as "cattle" using the LabelImg tool (Tzutalin, 2015). Then, the annotated images were used to count animals based on the Faster R-CNN object detection model (Ren et al., 2015). The quadcopter equipped with a high-resolution camera was flown at various heights (80, 90, 100, 110, and 120 m) over six different fields, to collect images at separate times of day (morning, noon, and afternoon). Three fields with drier and sandy soil, contained only white cattle, whereas the remaining three fields with reddish soil and greener pasture contained multiple color cattle (white, black, brown, red, and spotted/piebald).

### 3.1.20. Cattle_dataset

This dataset is prepared to classify individual cattle from their facial images using a DL classification model. Li et al. (2022) collected video data from 103 Simmental beef cattle from a farm using a digital color camera. The image resolution is 1,920 × 1,080, shot at 50 fps to reduce the motion causing blur. The videos (295 videos of 103 cattle), each 2 to 4 min long, were preprocessed to obtain 1000 frames per video. The invalid and similar frames were further filtered out. In this way, a total of 10,239 valid images (each cattle has 100 images except 4, which have fewer images) were made publicly available.

### 3.1.21. Aerial_pasture

This dataset was used to count the number of cattle in a pasture area by using multiple images (Shao et al., 2020), in which the detection results from multiple images were merged by reconstructing a 3-D model using the Structure from Motion (SfM) technique (Wu, 2013) to avoid duplicate counting. Shao et al. (2020) collected aerial images at 4,000 × 3,000 pixels resolution and curated two sub-datasets. Dataset 1 contains 656 aerial images taken on sunny and bright days from 50 m height, covering around 80 m × 60 m area, whereas Dataset 2 contains 14 images taken in a different area and season but with similar weather conditions. The dataset has manually annotated using a bounding box covering cattle objects with four types of data quality metadata (Normal, Truncated, Blurred, and Occluded). As a result, Dataset 1 has 1,886 labels from 212 individual targets with an average cattle size of 87 × 90 pixels, and Dataset 2 contains 62 annotations from 6 individual targets with 59 × 101 pixels.

### 3.1.22. Cattle_Video_Behaviors (CVB)

This dataset was used to identify eleven visually distinct and mutually exclusive cattle behaviors [grazing, walking, running, ruminating-standing, ruminating-lying, resting-standing, resting-lying, drinking, grooming, other (other than not specified behavior), and hidden)] (Zia et al., 2023). It is composed of 502 fifteen-second-long video clips. Each video was captured under natural lighting conditions on a grazing field at a rate of 30 fps, thus containing 450 frames per video. The cattle were detected and tracked using pre-trained YOLOv7 object detection models (Wang et al., 2022) for efficient annotation. Then the detected cattle were labeled using the Computer Vision Annotation Tool (CVAT) (https://www.cvat.ai/) (Sekachev et al., 2020). Eleven visually perceptible and mutually exclusive ethograms of cattle were labeled (Zia et al., 2023).

### 3.1.23. LShapeAnalyser-cow

This dataset was applied to measure and score the 3-D body shape of an animal by using the LShpaeAnalyser system developed by Zhang et al. (2023). The LshapeAnalyser-cow dataset (Zhang et al., 2023) consists of 203 dairy cows, which were captured during milking (to ensure little or no movement on cattle) using an RGB-D camera from the oblique rear view. The datasets do not have manually labeled key points; instead, the author employed an automatic tool, the DeepLabCut (Mathis et al., 2018), for landmark annotation (3 points on the upper body surface).

### *3.1.24. PCD Dataset*

The LiDAR sensor (O3D303, IFM Inc., Essen, Germany) generated 3D point cloud datasets (PCDs) from 100 beef cattle that were used for live body weight estimation is available in this dataset (Hou et al., 2023). The dataset is preprocessed by applying various filtering (filtering fusion of conditional, statistical, and voxel grid) followed by the RANdom SAmple Consensus (RANSAC) (Tong et al., 2015) and Euclidean distance clustering (Raguram et al., 2008) techniques, extracting clean 3-D contours of cattle. Thus, each voxel of the cattle body contains 3-D coordinates and label information. The cattle body surface is divided into four regions, as shown in Figure 5, using the Semantic-segmentation-editor tool (Hitachi Automotive and Industry Laboratory, 2018) where each voxel was labeled according to the regions.

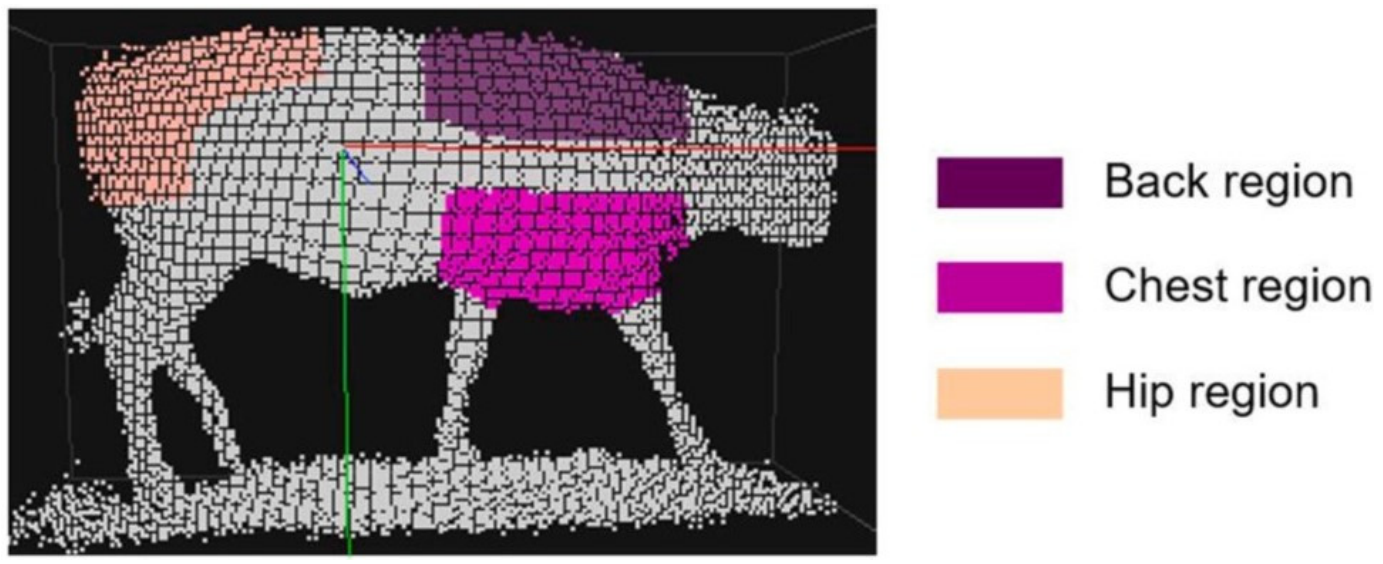

**Figure 5**. Segmentation of cattle body regions for labeling purposes; Back, Chest, and Hip regions for body measurements (Hou et al., 2023).

### *3.1.25. CattleEyeView*

Ong et al. (2023) collected a multi-purpose cow dataset for counting, detection, pose estimation, tracking, and semantic segmentation tasks by applying it to various deep-learning models. The dataset was recorded by RGB cameras over 4 months in the morning and evening. The acquired data were annotated with bounding boxes for the cow's body and head, unique tracking ID, actual cattle count, 24 key points for pose estimation, and segmentation masks. A total of 753 cattle instances were manually annotated with bounding boxes from 30,370 frames (14 video clips). Moreover, the dataset contains six diverse cattle breeds (Angus, Brahman, Charolais, Droughtmaster, Hereford, and Santa Gretrudis), having different body colors and patterns.

### *3.1.26. Dairy Cow*

This dataset was used to detect the cows and their daily poses (standing, lying, and walking) in the images of 10 distinct cows (Gong et al., 2022). Two surveillance infrared (IR) video cameras (Hikvision) configured with image resolution at 1920 × 1080 pixels and 25 frames per second frame rate were used to record videos of 10 animals. The LabelImg tool (Tzutalin, 2015) was used for animal annotation, and the Labelme tool ((Russell et al., 2008) for labeling the 16 lateral cow body key points (the head, left front leg root, right front leg root, left front knee, right front knee, left front hoof, right front hoof, left hind leg root, right hind leg root, left hind knee, right hind knee, left hind hoof, right hind hoof, neck, spine, and coccyx) in 1800 images. The author utilized YOLOv4 (Bochkovskiy et al., 2020) for background removal, followed by a pose classifier network for animal pose estimation (Cao et al., 2018).

### *3.1.27. Cows Frontal Face*

The muzzle pattern of an animal acts as a unique biological marker to distinguish one animal from others. This dataset consists of 2893 muzzle images from 459 cows. Three different color imaging devices (one digital camera and two smartphones) were utilized to capture frontal images of mixed cattle breeds in the early morning (Ahmed et al., 2024). The resultant data was annotated using the LabelImg tool (Tzutalin, 2015), and YOLO v7 (Wang et al., 2022) was trained to identify individual cow IDs amongst 459 distinct cows.

## 3.2. Swine Datasets

The second largest number of public datasets are associated with pigs. The bulk of the swine datasets were focused on animal posture and behavior identification. The lack of diverse public datasets is still a bottleneck in swine farming. The following subsections briefly describe each dataset, as summarized in Table 3, in terms of image acquisition, dataset size, animal annotation, and application.

### 3.2.1. PNPB Dataset

The PNPB (Pig Novelty Preference Behavior) dataset (Shirke et al., 2021a) was collected to study pig actions against novel objects. The dataset was prepared by capturing 20 RGB videos at a resolution of 1024 × 1024 pixels and a rate of 30 fps on an experimental farm and annotated at two levels (actions and key points of pigs). The raw videos were preprocessed and annotated in two levels: pig actions and key points. The pig's actions were recognized as “investigation” if the animal is around a novel object and “exploration” when the animal ignores the novel object. For the pig landmark label, bounding boxes on three key points (base of the tail, tip of the nose, and crown of the pig) in the frame were assigned on 668 frames from 9 video clips.

### 3.2.2. PigBehavior

In this dataset (Alameer et al., 2020), fifteen pigs (Landrace/Large White dams × synthetic sire line), housed under a commercial pig barn, were recorded using a ceiling RGB-D camera, at a microclimatic condition of 26.3 ºC ambient temperature and relative humidity 47% with the pig body weight of 50 kg. It was used to identify the pig’s various postures (standing, sitting, lateral lying, and sternal lying) and drinking behavior quantitatively and to track normal individual pig’s behavioral profiles under regular and limited resources (feed and water) conditions, using YOLOv2 (Redmon and Farhadi, 2016) and faster R-CNN (Ren et al., 2015) models followed by a tracking algorithm. This dataset contains 735,094 behavioural instances in 113,379 images for five ethogram types (that is, 105,132 standing pig annotations from 54,320 images, 23,801 Sitting from 22,417 images, 417,134 Lateral Lying from 105,199 images, 166,085 Sternal lying instances from 81,495 images, and 22,942 Drinking instances from 19,208 images).

### 3.2.3. PigFeeding-NNVBehavior

Alameer et al. (2020b) studied feeding and non-nutritive visits (NNV) behaviors of 15 pigs (Landrace/Large White synthetic sire line) of 9 to 14 weeks of age (33.6 – 51.0 kg weight). The video of pig behavioral activities was recorded using an RGB camera at a frame rate of 25 fps with a resolution of 640 × 360 pixels. A total of 34,375 images, divided into 7 categories of behaviors, were labeled with a number 1 or 2 representing the number of pigs engaged in a particular behavior (1 Pig Feeding: 2270 images, 1 Feeding and 1 NNV: 378 images, 1 NNV: 230 images, 2 Feeding: 27,736 images, 2 Feeding and 1 NNV: 933 images, 2 NNV: 2688 images and None: 140 images).

### 3.2.4. PigAgonisticBehavior

Han et al. (2023) developed a computer vision dataset to detect five interactive behaviors of grow-finish pigs. An RGB-D camera, installed above the feeders at a height of 2.44 m, was used to record pig social behaviors from two pens over six weeks. Six social groups (SG) with 10 pigs in each group were rotated through two test pens to produce a social interaction between pigs. The acquired video data was split into 30-frame short clips to ensure enough duration to complete one behavioral action without covering other actions. The raw footage was labeled into five behavioral classes (no-contact: NC, ear-to-body: EB, head-to-body: HB, levering: L, and mounting: M). Thus, a total of 15,679 videos (30 frames each) were formed after performing data augmentation on levering and mounting behavior video clips.

### 3.2.5. CountingPigs

The CountingPigs dataset was used for pig counting using a regression-based CNN counting model (Tian et al., 2019), which estimated the number of objects by summing the distributed density map after regressing the density map. Counting animals by spatial integration and local region analysis of density map could enhance the counting accuracy in case of shape distortion, occlusion, and different postures. The authors collected pig images from the internet and local farms. The resultant dataset consists of 3,401 images containing nearly 30,000 pigs, including 2,499 Internet-derived images and 902 images acquired at a commercial pig barn using a color camera. Then, the images were annotated with pig centroid coordinates and the number of pigs in an image.

**Table 3**. Public swine datasets.

| Dataset | Modality | Imaging environment | Dataset size | Annotation | Applications | URL |
|---|---|---|---|---|---|---|
| PNPB dataset (Shirke et al., 2021a) | RGB | Experimental farm | 20 videos of 5.5 minutes with 30 fps | Bounding box, Landmark | Behavior detection | https://drive.google.com/drive/folders/14XUYxM15NAI-zBrntrmQofhLv5otAw5b |
| PigBehavior (Alameer et al., 2020a) | RGB | Commercial pig barn | 7,35,094 instances from 1,13,379 images | Bounding box | Stress detection from posture | https://data.ncl.ac.uk/articles/dataset/Automated_recognition_of_postures_and_drinking_behaviour_for_the_detection_of_compromised_health_in_pigs/13042619 |
| PigFeeding-NNVBehavior (Alameer et al., 2020b) | RGB | Commercial pig barn | 34,370 images | Image-level | Behavior detection | https://data.ncl.ac.uk/articles/dataset/Automatic_recognition_of_feeding_and_foraging_behaviour_in_pigs_using_deep_learning/13084148 |
| PigAgonistcBehavior (Junjie Han et al., 2023) | RGB | Single space Feeding stall | 15,679 30-frame videos | Image-level | Behavior detection | https://osf.io/wa732/ |
| CountingPigs (Tian et al., 2019) | RGB | Internet and real pig barn | 3401 images from internet and real pig barn | Pig centroid coordinate | Counting by detection | https://github.com/xixiareone/counting-pigs |
| PigPosture (Riekert et al., 2020) | RGB | Commercial pig barn | 305 images | Bounding box | Posture detection | https://wi2.uni-hohenheim.de/fileadmin/einrichtungen/wi2/Publikationen/COMPAG_105391_Riekert_etal_2020.zip |
| Pig_Detection (Riekert et al., 2021) | RGB | Research center pig barn | 13,047 annotated pigs | Bounding box | Posture detection | https://github.com/majrie/pig_detection_dataset/raw/main/COMPAG_106213_Riekert_et_al_2021.zip |
| Pig_Behaviors (Bergamini et al., 2021) | RGB-D | Farm environment | 7200 annotated frames | Bounding box | Tracking by detection | https://drive.google.com/drive/folders/1C_wABDzfpdaRykVHoWSN8vAaLXs8Yaxn. |
| Multi-camera-pig-tracking (Shirke et al., 2021b) | RGB | Commercial barn | 429 images for detection 1200 frames for tracking | bounding box with pig ID | Tracking by detection | https://drive.google.com/drive/folders/1E2wW2aRENgy_TqlzfICn58ahbTHVIaK6 |
| Pig-multi-part-detection (Psota et al., 2019) | RGB and IR | Commercial pig barn | 2,000 images | Landmark (left and right ear, tail, and shoulder) | Pose estimation | http://psrg.unl.edu/Projects/Details/12-Animal-Tracking |
| Pig_tracking (Psota et al., 2020) | RGB video | Commercial pig barn | 15 videos each of 30 minutes at 5 fps | Landmark (shoulder and tail) with pig ID | Tracking by detection | http://psrg.unl.edu/Projects/Details/12-Animal-Tracking |
| Pig-detection&tracking (Wutke et al., 2021) | Gray | Research farm | 49 raw videos | Landmark (left and right ear, tail, and shoulder) | Tracking by detection | https://github.com/MartinWut/Supp_DetAnIn |
| PigPostureAcitvity (Bhujel et al., 2021) | RGB | Experimental pig barn | 668 images | Bounding box | Posture behavior detection | https://drive.google.com/file/d/1DmkR5AyysQkFbMEwjPjJnnNVyGvtsu9H/view |
| PigTrace (Tangirala et al., 2021) | RGB | Commercial farm | 540 frames from 30 videos | Semantic segmentation with pig ID | Behavior detection | https://drive.google.com/file/d/1s-bCnABh2Hef5l5OxydcY-tkPbrUGSjj/view |
| LShapeAnalyser Dataset (Zhang et al., 2023) | RGB-D | Indoor Cage | 201 pigs | Landmark | Body measurement | https://gitee.com/kznd/lshape-analyser/tree/master/Dataset/pig |
| DifferentStagesPig (Pan et al., 2023) | RGB | Commercial farms | 319 images and 17 videos | Image-level | Healthy and unhealthy detection | https://data.mendeley.com/datasets/vd5vmgr8kg https://data.mendeley.com/datasets/jy6hngx7df |
| Aggressive-Behavior-Recognition (Gao et al., 2023) | RGB | Commercial farm | 541 aggressive and 565 non-aggressive behavior video clips | Video-level | Behavior detection | https://github.com/IPCLab-NEAU/Aggressive-Behavior-Recognition |

### 3.2.6. PigPosture

Riekert et al. (2020) collected videos from 18 pens having different pig ages, using 21 cameras (18 from high angle and 3 from top view perspectives). Each pen contains a different number of pigs (2 fattening pig pens contain 18 pigs each, one fattening pen has 28 pigs, one piglet pen contains 24 pigs, and another piglet pen contains 48 pigs). Two types of RGB video cameras were used for acquiring images at a rate of 10 fps. A set of randomly selected 305 images taken from video clips were annotated with bounding boxes into two classes of postures (lying vs. not lying). Since most of the time pigs remained lying, an imbalance exists between lying and not lying classes, where among the total 7,277 annotations, there are 5077 lying and 2,200 not lying pigs.

### 3.2.7. PigDetection

This dataset contains both the PigPosture (Riekert et al., 2020) and additional IR video frames collected by Riekert et al. (2021). The dataset was used as a benchmark for evaluating the performance of various DL models in detecting pig's position and postures (lying and not lying). The data was collected from a research center pig barn having 18 fattening pigs with an equal number of castrated male and female pigs. The pig posture was manually labeled into two classes (lying and not lying), producing 13,047 bounding boxes. As in PigPosture (Riekert et al., 2020), the intra-class bounding boxes are largely imbalanced in this dataset since the pigs remained inactive most of the time, with 2,997 annotations of not lying vs 10,050 annotations of lying.

### 3.2.8. Pig_Behaviors

This dataset was used for the detection, tracking, and long-term behavior analysis of pigs (Bergamini et al., 2021). The authors used YOLOv3 for animal detection, a tracking-by-detection algorithm for tracking, and behavioral analysis for finding behavioral changes like standing-to-walking using tracking-by-detection results with temporal information. The dataset is composed of 8 growing pigs from a single pigpen collected for six weeks. An RGB-D camera, placed at 2.5 m from the ground, was used to capture both RGB and depth images at 6 fps with a resolution of 1280 × 720 pixels during daytime. The video was split into 5-minute video clips (1,800 frames), thus creating 1905 clips. Although the authors collected 3.4 million frames, the public dataset consists of only 7,200 images from 12 clips, annotated with bounding boxes using five pig behavior labels (lie, move, eat, drink, and stand) and a unique identity for each pig.

### 3.2.9. Multi-camera-pig-tracking

This dataset sets a new benchmark for detecting and tracking pigs in the entire pen area captured from multiple cameras. Multi-camera frames were applied to pig detection with YOLOv4 (Bochkovskiy et al., 2020), and the detected animals were assigned using local IDs by a DeepSORT algorithm (Wojke et al., 2018). A homography estimation technique was employed for uniquely assigning the pig's global IDs. The video frames were collected from two grow-finish pens containing 16-17 pigs per pen using three wide-angle RGB cameras (one mounted on the ceiling at 4 m height and two angled cameras at 2.2 m) (Shirke et al., 2021b). In total, 429 images were annotated using the VGG Image Annotator (Dutta and Zisserman, 2019), and 1200 annotated frames (one frame per second extracted from five two-minute videos) with assigned global pig IDs were made available for pig tracking.

### 3.2.10. Pig-multipart-detection

The purpose of this dataset was to identify the location and orientation of all the pigs visually included in a frame (Psota et al., 2019). The author adopted a top-down approach where the key body parts (tail, shoulder, left ear, and right ear) of the animal were identified first and then reconstructed the whole-animal instances. Various body parts of 5.5-month-old pigs were imaged with RGB and IR cameras from varied heights of 2.5 to 6 m facing vertically down, at a group-housed pig barn in 17 different locations. A total of 2,000 images were annotated for different body parts of each visible pig (left ear: red, right ear: green, shoulder: blue, and tail: yellow). The dataset gathers a wide range of animal poses since each frame was extracted from a long-span video recorded for more than a week at each location.

### 3.2.11. Pig-tracking

This dataset (Psota et al., 2020) was curated for pig tracking at the individual level after detection. For this purpose, a total of 15 videos, each 30 minutes long, with a resolution of 2688 × 1520 and 5 fps, were captured for diverse pig activities at a commercial pig barn. The dataset was annotated with the pig's shoulder and tail key points along with a physical ear tag ID, consisting of five videos from nursery pigs (3 – 10 weeks old), five from the early finisher phase (11-18 weeks old), and the last five videos from the finisher phase (19-26 weeks old). Moreover, in each phase video, 3 videos were collected during the daytime with the light on, and the remaining two videos were captured at night using an IR camera by illuminating the pig barn with IR flood lights. The details of activity level, number of pigs, and video recording time that the dataset contains are presented in Table 4.

**Table 4**. Details of the video data of the Pig-tracking dataset (Psota et al., 2020), where "H", "M", and "L" means the High, medium, and low activity level of pigs in the video.

| Video# | Nursery | | | | | Early Finisher | | | | | Late Finisher | | | | |
|---|---|---|---|---|---|---|---|---|---|---|---|---|---|---|---|
| | 1 | 2 | 3 | 4 | 5 | 6 | 7 | 8 | 9 | 10 | 11 | 12 | 13 | 14 | 15 |
| Day | √ | √ | √ | | | √ | √ | √ | | | √ | √ | √ | | |
| Night | | | | √ | √ | | | | √ | √ | | | | √ | √ |
| No. of Pig | 16 | 16 | 15 | 16 | 16 | 7 | 15 | 7 | 8 | 8 | 16 | 14 | 12 | 14 | 13 |
| Activity Level | H | M | L | M | L | H | M | L | M | L | H | M | L | M | L |

### *3.2.12. Pig-detection&tracking*

The social interactions of pigs were studied by detecting the location and orientation of animals and tracking their movement trajectories over a period (Wutke et al., 2021). The authors utilized the video data collected by Lange et al. (2020) while studying tail lesions and losses of docked and undocked pigs reared and farrowed in different environments by Gentz et al. (2020) during the studying of post-weaning stress due to different farrowing and rearing systems. An RGB camera, fixed at 3 m height from the ground, was used to record the videos at a rate of 10 fps and 1,280 × 800 pixels resolution. The acquired video data were processed to extract frames and converted to grayscale, resulting in a total of 2457 images annotated with pig's ears, shoulder, and tail landmarks. The spatial information of the four different body parts (left ear, right ear, shoulder, and tail) and three connection lines (shoulder-tail, shoulder-left ear, and shoulder-right ear) of a pig were stored as a binary image in a separate image channel, as shown in Figure 6.

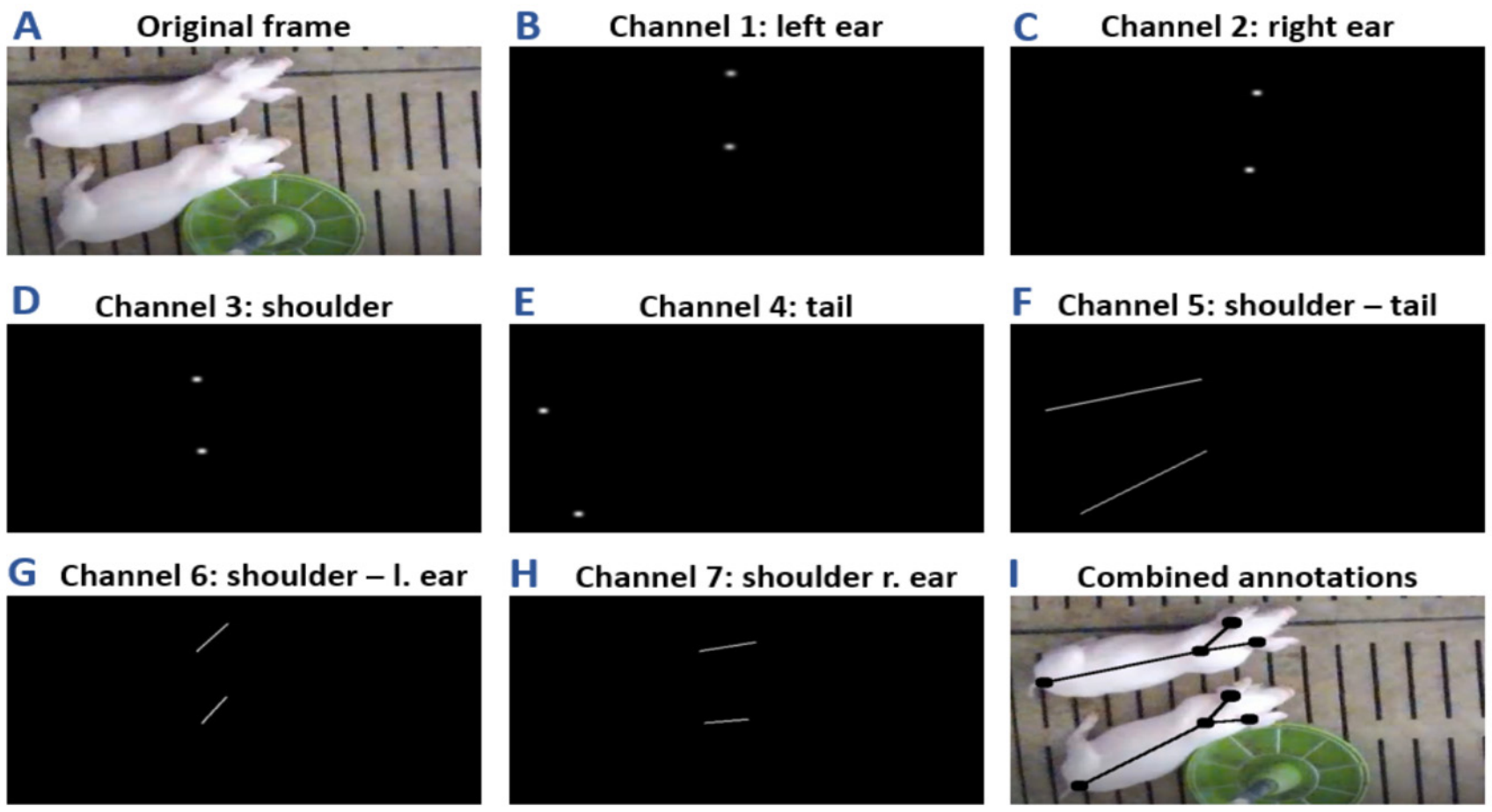

Figure 6. Dataset annotation: (A) The original image, (B–H) The corresponding ground truth annotations containing the positional body part information, which are used for the training process, (I) The original image combined with the ground truth annotations (Wutke et al., 2021).

### *3.2.13. PigPostureActivity*

Various pig postures, such as sternal lying, lateral lying, and standing, were identified and tracked in a compromised environment using an end-to-end tracking-by-detection model (Bhujel et al., 2021). The videos of the dataset were taken using a top-view camera installed at an experimental pig barn housing 5 pigs at 30 fps with a resolution of 1,920 × 1,080 pixels, which was later downsized to 640 × 640 pixels and filtered out for redundant frames. The dataset was prepared by manually labeling pig postures (lateral lying, sternal lying, and standing) using the computer vision annotation tool (CVAT, https://www.cvat.ai/), and walking activity was detected by analyzing the standing pig's position within consecutive frames. A total of 668 testing frames with label information, sample videos of pig posture identification from the Faster R-CNN and YOLOv4 models, and a pig tracking video from the tracking algorithm were available in the dataset.

#### *3.2.14. PigTrace*

Tangirala et al. (2021) developed a STARFORMER system, a group-housed pig behavior monitoring system accomplishing pig's instance segmentation, tracking, action-recognition, and re-tracking (STAR) tasks using Transformer (FORMER) architecture. The PigTrace dataset (Tangirala et al., 2021) was used as a benchmark dataset for evaluating the system performance. It consists of 540 frames associated with 30 videos of 5 seconds collected from five commercial farms. A set of 210 frames from 7 video sequences at a rate of 6 fps and 345 frames from 23 videos at a 3-fps rate were annotated for pig eating, drinking, standing, and walking activities with a unique ID and semantic masking on each pig.

#### *3.2.15. LShapeAnalyzer-pig*

The 3D images of local body parts of the animal were applied to the LShapeAnalyser tool developed by Zhang et al. (2023) with the objective of scoring body condition quantitatively by reconstructing the livestock body shape from key points. The dataset was built from the top-view of 201 pigs taken in a cage one by one using a a consumer-grade RGB-D camera. For body shape analysis, landmarks of the animal body surface were extracted using an automatic key point detection algorithm (DeepLabCut) (Mathis et al., 2018).

#### *3.2.16. DifferentStagesPig*

The objective of this dataset was to implement a sorting system according to the pig status (disease, breeding, growth, and fattening) (Pan et al., 2023). Various morphological features of pigs, such as marks on skin, ear, eye, nose, tail, feed, body weight, etc., were extracted and analyzed by a DL network to identify the actual status of individuals for sorting them automatically. The dataset covers a wide range of pig images taken at different stages. A total of 63 images and a video of 50-day-old piglets, 139 images and 3 videos of 70-80-day-old piglets, 91 images and 5 videos of fattening pigs, 26 images and 4 videos of sows, and 4 videos of male breeding pigs were gathered. All the data were captured from 2021 to 2023 using front and top-view cameras.

#### *3.2.17. Aggressive-Behavior-Recognition*

Gao et al. (2023) constructed aggressive and non-aggressive behaviors of 35-42-day-old Large white x Landrace piglets. A hybrid model using CNN (for spatial feature extractor) and a gated recurrent unit (GRU) (temporal feature extractor) were used to identify aggressive social behaviors (sniffing and light pushing followed by biting and bumping) of pigs. The aggressive behaviors were recorded using a top-view RGB camera after mixing the different groups of piglets in a pen. Biting (tail biting, ear biting, and body biting), knocking (head and body knocking), trampling, and chasing actions were considered aggressive behavior while the mounting, playing, lying, feeding, and drinking activities were taken as non-aggressive. The resultant dataset has 541 video episodes of aggressive behaviors and 565 video episodes of non-aggressive behaviors with a duration of 3 seconds each.

### 3.3. Poultry Datasets

Poultry datasets (Table 6) typically consist of a collection of images or videos related to poultry farming or the study of poultry birds. These datasets are commonly used in machine learning tasks for various applications, such as poultry disease detection, counting, and behavior analysis. The difficulty of poultry monitoring lies in the complex background and variations in illumination in a real pen environment. Though poultry production represents an important sector of agriculture and food supply, the computer vision datasets for them are comparatively scanty. The following subsections present those publicly available datasets.

#### *3.3.1. ChickenGender Dataset*

Yao et al. (2020) prepared a chicken gender database for gender classification by DL models. This dataset comprises 1,000 images of chickens extracted automatically from 800 flock images by a trained YOLOv3 detector. Each chicken image was manually labeled either as rooster or hen class. A digital color camera with a resolution of 2992 × 2000 pixels was employed to collect images. Although the images were taken from different distances with vertical and horizontal angles, the wide-distance top-down images dominate the dataset. The dataset contains chicken gender images in various natural behaviors like eating, drinking water, and waving wings, which provides diversity.

### 3.3.2. Chicken_Poo

This dataset was used to relate the health status of chickens by using chicken fecal images (Aworinde et al., 2023). It gives information about the healthy or unhealthy condition of broilers and layers, but not what health issue a particular chicken has. The healthy and unhealthy chicken fecal images are stored separately in the corresponding folders for classification purposes. A total of 14,618 processed and labeled images (7991 healthy and 6627 unhealthy) were made publicly available, which were derived from images captured by an RGB camera at different times of the day (morning, afternoon, night, and others).

### 3.3.3. Poultry_Disease

The *Poultry_Disease* dataset contains fecal images of sick (Coccidiosis, Newcastle, and Salmonella) and healthy chickens (Machuve et al., 2022). The images were acquired from layers, cross, and indigenous breeds of chicken breeds periodically for a year, with an ordinary smartphone installed with the Open Data Kit (ODK) tool (Hartung et al., 2010). The images were preprocessed and stored annotated at the image level for classification purposes. Moreover, the dataset was labeled using bounding boxes and masks to fit into the object detection and semantic segmentation model using LabelImg (Tzutalin, 2015) and LabelMe (Russell et al., 2008) tools. The resultant repository constitutes two subsets: Dataset 1 has 1255 fecal images with a Polymerase Chain Reaction (PCR) report, and Dataset 2 contains 6,812 fecal images from four classes of chicken health conditions. The detailed composition of the two subsets is presented in Table 5 (Machuve et al., 2022). The authors evaluated DL models (VGG16, InceptionV3, MobileNetv2, and Xception) on the dataset for classifying the fecal images of sick and healthy chickens.

**Table 5**. Data distribution of the chicken disease dataset (Machuve et al., 2022).

| Class-fecal images | Laboratory labeled (Dataset 1) | Farm labeled (Dataset 2) |
|---|---|---|
| Healthy | 347 | 2057 |
| Salmonella | 349 | 2276 |
| Coccidiosis | 373 | 2103 |
| Newcastle | 186 | 376 |
| Total | 1255 | 6812 |

### 3.3.4. GalliformeSpectra

Himel and Islam (2024) published the *GalliformeSpectra* dataset for ten hen breeds, from which the CV community could leverage to design and test algorithms for chicken breed classification. The dataset contains 1010 images, with 101 images for each of the 10 hen breeds, which were captured by handheld color cameras and smartphones, from farms located in various countries with diverse backgrounds.

### 3.3.5. Broiler Dataset

Elmessery et al. (2023) developed and tested thermal and visual images of chickens for various pathological phenomena such as lethargic chickens, slipped tendons, diseased eyes, stressed (beaks open), and pendulous crop, including healthy broilers. The dataset was collected via smartphones and a thermal camera from 600 Cobb Avian, where 48 male and female broilers of different ages were grown. The resultant dataset contains 10,000 RGB and thermal images alongside 50,000 bounding box instances.

**Table 6**. Public poultry datasets.

| Dataset | Modality | Imaging environment | Dataset size | Annotation | Applications | URL |
|---|---|---|---|---|---|---|
| ChickenGender (Yao et al., 2020) | RGB | Commercial poultry farm | 800 chicken flock and 1,000 single chickens | Bounding box | Chicken gender detection | https://drive.google.com/drive/folders/1eGq8dWGL0I3rW2B9eJ_casH0_D3x7R73 |
| Chicken_Poo (Aworinde et al., 2023) | RGB | University's & community poultry farm | 14,618 healthy and non-healthy chicken poo | Bounding box | Disease detection | https://data.mendeley.com/datasets/8pnbzpt2k9/1 |
| Poultry_Disease (Machuve et al., 2022) | RGB | Small to medium poultry farm | Dataset1: 1255 fecal images, Dataset 2: 6812 from 4 categories | Image-level | Disease detection | https://zenodo.org/record/4628934#.YtDNzOxBy1u https://zenodo.org/record/5801834#.YtDN9-xBy1t |
| GalliformeSpectra (Himel and Islam, 2024) | RGB | Chicken farm | 1010 images | Image-level | Chicken breed detection | https://data.mendeley.com/datasets/nk3zbvd5h8/1 |
| Broiler Dataset (Elmessery et al., 2023) | RGB and Thermal | Research farm | 10,000 images with 50,000 annotations | Image-level | Disease detection | https://drive.google.com/drive/folders/1jj9LKL0d1YDyDez8xrmKWRWd3psFoeZ2?usp=sharing |

## 3.4. Datasets for Other Livestock Animals

Other than the above-mentioned datasets, nine livestock CV datasets identified are related to other livestock species (Table 7). Amongst them, seven datasets are related to sheep and goat monitoring, whereas two datasets belong to buffalo and horse-related studies.

### *3.4.1. SheepBreed*

This dataset (Abu Jwade et al., 2019) was captured from 160 sheep of four different breeds in a commercial farm environment using an RGB camera. The camera recorded video at 24 fps with 1,920 x 1,080 resolution. The unobstructed sheep face was captured using four cameras placed in multiple positions. The authors compiled around 400 images from each sheep breed to avoid inter-class data imbalance, which resulted in a total of 1642 images. The acquired images were annotated manually at the image level by labeling each image according to sheep breed.

### *3.4.2. SheepBase Dataset*

This dataset contains 6,559 sheep face images of two sheep breeds captured in the natural environment (Xue et al., 2021). The SheepBase dataset was developed after several preprocessing operations (cropping and aligning to reduce facial posture and angle influence). In this way, three sub-datasets (the sheep face target detection, the sheep face key point detection, and the sheep face recognition) were constructed, which were used for identifying the biological signature of sheep from their faces.

### *3.4.3. SheepActivity Dataset*

This dataset contains various sheep activities, such as grazing, standing, walking, running, and sitting, recorded in a natural environment (Kelly et al., 2024), with a total of 149,327 frames from 417 videos shot using smartphones from various angles (sagittal and side). Additionally, 8 videos belonging to extra activities are also included in the dataset. Since the authors did not apply any computer vision algorithm to this dataset, its quality and usability are yet to be tested.

### *3.4.4. LEsheepWeight*

The LEsheepWeight dataset (He et al., 2023) was made by recording side-view 3D videos of 726 sheep with ages ranging from 4 to 28 months. An RGB-D camera was used for the data collection. After preprocessing, a total of 6,373 RGB frames with depth information and the corresponding weight (measured manually) are provided in the dataset. The authors used the dataset with a LiteHRNet (Lightweight High-Resolution Network) to estimate the live weight of sheep (He et al., 2023).

### *3.4.5. GoatImage*

The GoatImage dataset (Billah et al.,2022) contains 3,278 images of goat faces, with 1967 images collected from open sources and 1,311 images of 10 individual goats captured from a local farm using an RGB camera. The dataset was further divided into two sub-datasets. The dataset 1 comprised 1,680 open-source images labeled with bounding boxes for facial landmarks (face, eyes, nose, and ears), and similarly, the dataset 2 contains images collected from a farm for face recognition. The authors automatically labeled 13,761 instances (3078, 3326, 2586, and 4771 bounding boxes for face, eye, nose, and ear, respectively) from the images in dataset 1. The datasets were applied to YOLOv4 to detect the goat's face and facial landmarks (Billah et al.,2022).

### *3.4.6. Drone-goat-detection*

This dataset (Vayssade et al., 2019) consists of images taken from two different flocks of goats, using a drone with an RGB camera, flown at a height of 50 m from the ground. A set of 517 images was annotated to 8,830 goat-centroids and bounding boxes with corresponding goat activities (sleeping and grazing). A goat activity detection system using various image processing and classifier algorithms was developed and validated with this dataset.

### *3.4.7. CherryChevre*

Vayssade et al. (2023) collected 297 images of three breeds of goats from diverse lighting and environmental conditions using multiple RGB cameras at various times over two years. The dataset comprises 6160 images that are sub-grouped into five subsets for annotation. A total of 35381 unique bounding box annotations made using VGG Image Annotator (VIA) (Dutta and Zisserman, 2019) are available in this dataset (Vayssade et al, 2023). This dataset was tested for goat detection using YOLOv8 with images taken at oblique camera angles.

### 3.4.8. Buffalo-Pak

This dataset consists of both labeled and raw images of three Buffalo breeds (Pan et al., 2022). A total of 325 RGB images from the three breeds were collected and annotated at the image level according to the breed. The authors also performed augmentations to increase the data volume to 450 images and applied it to a self-activated CNN for classification.

### 3.4.9. Horse-10

This dataset (Mathis et al., 2019) comprises 8,114 frames of 30 individual horses captured by an RGB camera at 60 fps with a frame resolution of 1,920 × 1,080 pixels, with images annotated for 22 anatomical landmarks. They used the DeepLabCut labelling tool (Mathis et al., 2018) to label different body parts. The availability of more than 200 postures of each horse in this dataset enriched the horse pose detection and its application in real-world scenarios.

**Table 7**. Public datasets of other livestock species.

| Dataset | Modality | Imaging environment | Dataset size | Annotation | Applications | URL |
|---|---|---|---|---|---|---|
| SheepBreed (Abu Jwade et al., 2019) | RGB | Commercial farm | 1,642 images from 4 breeds | Image-level | Sheep breed detection | https://data.mendeley.com/datasets/64gkbz8bdb/2 |
| SheepBase (Xue et al., 2021) | RGB | Natural environment | 6,559 sheep face images | Bounding box | Sheep face detection | https://pan.baidu.com/s/ 1HgNdEYqAz2SXpEbrmEb8UA (extraction code: zrcp) |
| SheepActivity (Kelly et al., 2024) | RGB | Natural environment | 149,327 frames from 417 videos | Image-level | Activity detection | https://data.mendeley.com/datasets/w65pvb84dg/1 https://data.mendeley.com/datasets/h5ppwx6fn4/1 |
| LEsheepWeight (He et al., 2023) | RGB-D | Sheep Breeder | 6,373 images from 726 Sheep | Corresponding live weight | Live weight estimation | https://pan.baidu.com/s/1lkF50WdG6vWCnj1TAw_LjA (code: 9hks) |
| GoatImage (Billah et al., 2022) | RGB | Open source and farm | 3,278 images | Bounding box on key points | Face detection | https://data.mendeley.com/datasets/4skwhnrscr/2 |
| Drone-goat-detection (Vayssade et al., 2019) | RGB | Goat grazing fields | 517 images with 8,830 goat-centroids | Bounding box with Goat-centroid | Activity detection | https://gitlab.com/inra-urz/drone-goat-detection |
| CherryChevre (Vayssade et al., 2023) | RGB | Farms and grazing pasture | 6160 annotated images | Bounding box | Activity detection | https://doi.org/10.57745/QEZBNA |
| Buffalo-Pak (Rauf, HT and Lali, MIU, 2021) | RGB | Not mentioned | 325 images | Image-level | Buffalo breed detection | https://data.mendeley.com/datasets/vdgnxsm692/2 |
| Horse-10 (Mathis et al., 2019) | RGB | Outdoor environment | 8,114 frames from 30 horse | Landmark (22 | Pose estimation | http://www.mackenziemathislab.org/horse10 |

# 4. Discussion

The application of CV and DL methods in the livestock field has proliferated in the last decade. However, the dearth of publicly available, high-quality livestock datasets has created a bottleneck to accelerating CV and DL developments and practical implementation in livestock farming. The existing public datasets are generally of small or medium size in volume, which may not be sufficient for deep architecture development and complex domain implementation. Hence, the following subsections unfold various dimensions of livestock CV with identified issues, potential solutions, and achievements.

## 4.1. Application-wise Dataset Taxonomy

The publicly available computer vision datasets for livestock monitoring are dominated by animal detection, followed by posture and behavior detection. Figure 7 depicts the statistics of public datasets categorized based on applications in livestock farming.

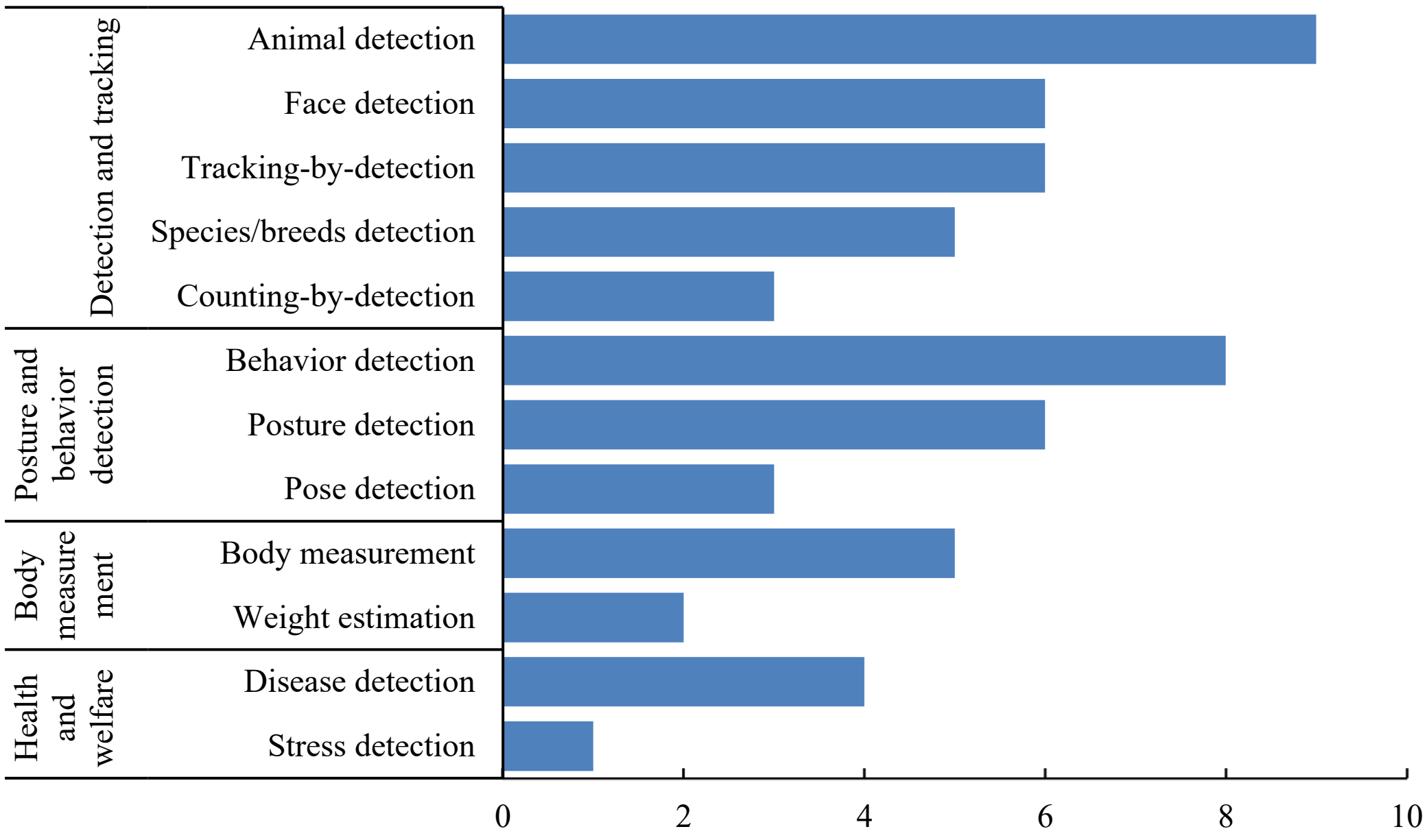

**Figure 7**. Number of public livestock datasets according to application type.

The application of livestock datasets for understanding the physiological characteristics of animals is crucial for health and welfare assessment. The RGB image provides spatial information, allowing algorithms to localize and detect animals and their pose, like lying, sitting, etc. However, mere 2D visual information may not be sufficient to precisely identify various behavioral activities. Livestock video data caters to flexibility in feature extractions with spatial and temporal information that play a pivotal role in detecting animals' social behaviors. Although there are a notable number of public datasets regarding animal detection from the top view, inadequate datasets exist for livestock body measurements, tracking, pose estimation, and health detection and monitoring. To address the dataset gaps or deficiencies, the synergy of spatial and temporal data becomes essential, which would empower practitioners in livestock farming to generate data-driven decisions, enhancing animal welfare, and production.

## 4.2. Imaging Modalities

The choice of imaging modality plays a pivotal role in the development of robust and versatile datasets. Figure 8 shows a comparison of different imaging modalities for livestock species. Datasets catering to PLF applications encompass diverse modalities that offer unique advantages and insights into livestock monitoring and assessment. RGB cameras are the most widely used for livestock imagery, because of their simplicity, accessibility, affordability, and efficiency. However, 3D cameras and point clouds have gained popularity for volumetric measurements of animals, and weight and pose estimation. On the other hand, the integration of multiple modalities holds promise for enhanced performance of PLF systems, ultimately benefiting production efficiency and farm management practices. Currently, there is still limited research into the development of multimodal vision systems in PLF, warranting future efforts.

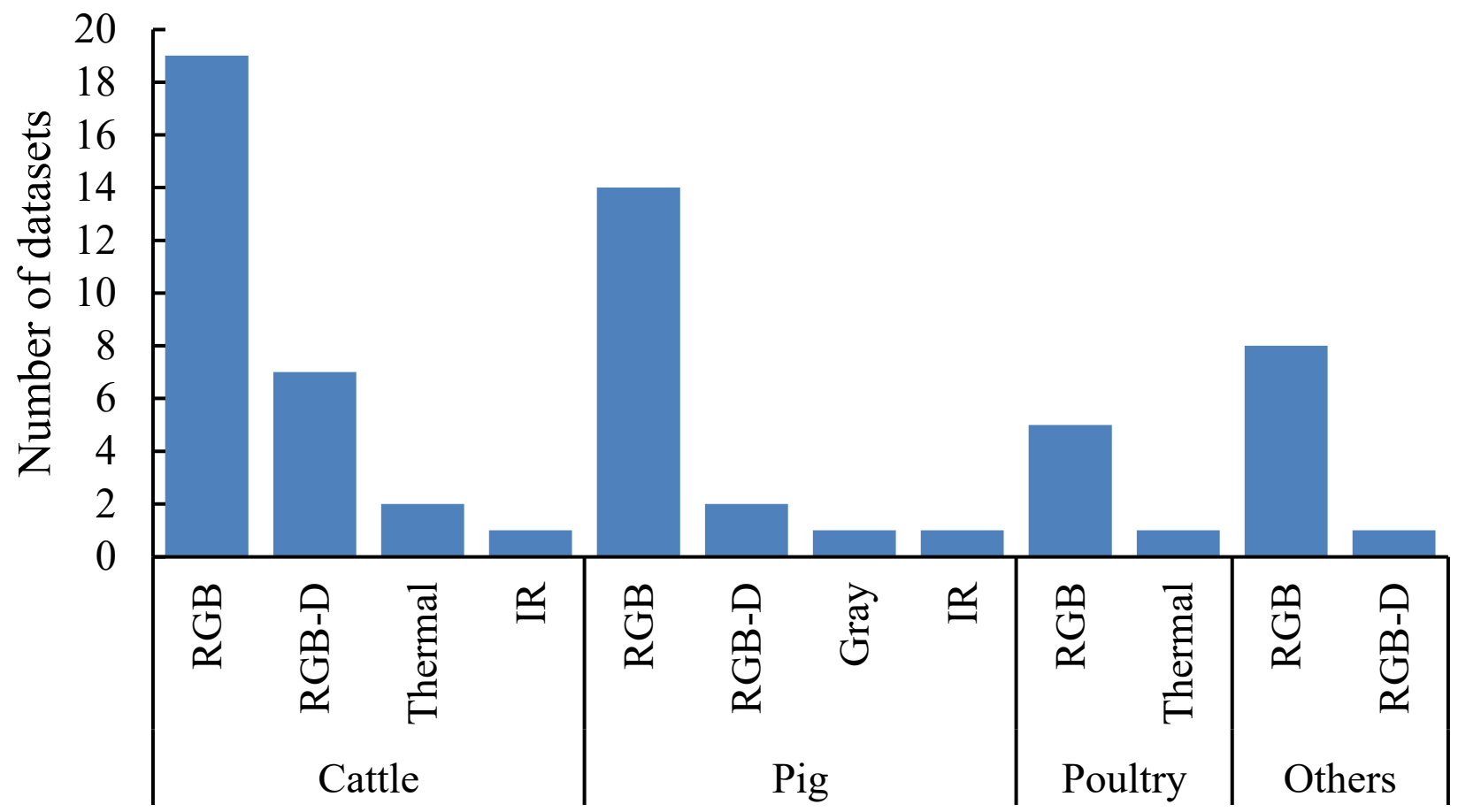

**Figure 8**. Number of public datasets of different livestock species according to imaging modality.

### 4.3. Challenges and Prospects of Livestock Dataset Curation in PLF

The preparation of image datasets for livestock farming applications is a vital endeavor that underpins the advancement of precision agriculture, enabling the implementation of state-of-the-art CV and DL techniques for the specific demands of livestock monitoring and analysis. In the following subsections, we delve into the general challenges encountered in the dataset curation process, limitations or constraints in existing public datasets, and the prospects of future research in PLF.

#### *4.3.1. Individual or group-housed animal detection and tracking*

The effective detection and tracking of animal stem from the latest advancements in CV algorithms, imaging technology, and adequate training data. Despite significant progress made in developing livestock detection techniques over the past decade, individual detection and tracking of spatially invariant animals like pigs and broilers in complex environments while ensuring high reliability and accuracy is still a challenging task for both researchers and farmers. Some studies utilized unique biological markers like face, iris, muzzle pattern, etc., to detect livestock individually from the herd. However, automatic collection and preprocessing of facial or muzzle data are time-consuming tasks, demanding many steps while preparing such a unique biomarker dataset. Li *et al*. (2022) reported nearly 85% of captured frames were invalid and filtered out while preparing a facial dataset from the raw data captured by a surveillance camera.

In addition, even if a considerable amount of open CV datasets is available for animal detection either at the individual or group level, there are few top-notch annotated datasets with ample metadata documenting farm environments, lighting conditions and amount, background type and features, animal posture, size, and density, occluded and overlapped animal, frame coverage area, etc. Moreover, a chunk of datasets having top-view scenes focused on visually distinctive individual animals like Holstein dairy cows, which have perceptible differences in color and dorsal pattern. In terms of animal tracking, numerous tracking algorithms showed promising results when the targeted object remained within the camera vision and the detection algorithm worked well. However, tracking multiple animals in densely populated commercial stockings in long video recordings is still a daunting task for computer vision researchers (Alameer *et al*., 2020a). Nonetheless, multi-view images with meticulously designed algorithms could potentially overcome such challenges.

#### *4.3.2. Animal posture and behavioral detection*

Among the most critical applications in PLF is posture and behavioral monitoring of animals, since it provides insights into animal health and welfare. Animals express their internal and external feelings through their posture and behavior. The majority of public datasets deal with particular animal species, behavior, and contextual information. For instance, around 60% of posture and behavioral detection datasets were acquired for pigs. The livestock animal poses a wide variety of postures and behavioral activities. Some postures and activities (lying, feeding, standing, grazing, etc.) can be detected considering only spatial features. However, certain behaviors, such as social interactions, walking, running, etc., are complex and challenging to identify from single frames (Gao et al., 2023; Liu et al., 2020). Thus, spatio-temporal features would be beneficial for precisely detecting posture and behavior activities. Moreover, contextual information, for instance, social dynamics, individual characteristics, health status, transitional or mixed behaviors, along with farm and animal-related information, could enhance the detection and generalizability of the model.

#### *4.3.3. Health and wellbeing monitoring*

Changes in animal posture, pose, and activity are the fundamental signs while assessing animal health and well-being. Physiological changes such as lesions, rashes, scars, etc., alongside the rise in body temperature, provide important clues in the manual inspection of livestock health and welfare. Moreover, faecal images were utilized in literature, especially in poultry industries, to detect various diseases. The non-invasive and contactless nature of CV technology permits safe and stressless livestock monitoring, preventing the spread of zoonotic diseases. Although it is agreed that the study of postures and activities of animals can be associated with their actual health and well-being status, meager datasets are directly acquired from genuinely disease-affected cases (Aworinde et al., 2023; Machuve et al., 2022). Most of the reported CV studies are analogy-type in terms of health detection, with only four datasets available that were acquired from diseased animals, accentuating the need for datasets on quantitative measurements of animal health conditions and welfare.

#### *4.3.4. Body shape and weight estimation*

Live body measurement and weight estimation of animals have multifaceted benefits in livestock farm management. It allows farmers to manage resources (feed, water, vaccination, etc.), animals (sorting and segregation), health and growth monitoring. Recently, the livestock dimension measurement and weight estimation tasks exploited 3D vision techniques with both spatial and depth information, attaining remarkable performance in 3D body reconstruction (Lu et al., 2022). Various techniques have been applied in reconstructing the 3D shape of animals, such as employing multi-angle 2D cameras (Thapar et al., 2023), laser scanning (Time-of-flight) (Omotara et al., 2023), LiDAR (Light Detection and Ranging), and depth camera (Gebreyesus et al., 2023; Ruchay et al., 2022a). Although 3D imaging technology may be susceptible to environmental factors and quick motion of objects, the latest improvements in hardware and software open up opportunities for enhanced model performance.

In body dimension measurement, both top-down and bottom-up approaches have been researched. The bottom-up approach first detects the anatomical key points of an animal and reconstructs a skeleton shape before identifying body parameters. Thanks to well-annotated body landmarks used for model training, the key points detection accuracy increases, thereby producing a promising body measurement accuracy (Yik et al., 2020). In contrast, the top-down approach first localizes and extracts animal contours and identifies animal body surface landmarks for measurements. Thus, the accuracy of the model using the top-down approach tends to be lower than the bottom-up approach, while the model throughput is higher than the latter as it can detect and analyze multiple animals in a single frame. Finally, a regression model is used to estimate the live body weight of the animals using phenotypic data. The major challenges in body measurement and weight estimation of livestock are limited models to deal directly with 3D data, an obvious lack of quality public 3D datasets, and noise-prone systems (Ma et al., 2024). Using environmentally robust sensors (stereo camera) and supplementary information, like animal position and postures, farm ambient data, object shape information (deformed or regular), etc., could address these challenges.

#### *4.3.5. Imaging modalities*

Although RGB or color imaging are the most widely used for CV tasks, they may not reliably deliver human-level accuracy in complex scenarios (social behavior detection and body measurement). Recently, 3D imaging has drawn significant attention due to its capability to provide additional depth or ranging information that can be beneficial for a variety of applications (e.g., body condition assessment). However, some challenges surround 3D imagery acquisition and processing. The imaging process is often sensitive to environmental factors (light, temperature, humidity, etc.) and object motion, producing noise and artifacts (Zhang et al., 2023b). 3D image data demands more preprocessing steps, sophisticated analyzing tools, and skilled manpower. The depth resolution and accuracy of 3D imaging may be limited, depending on sensor technology, analyzing algorithms, and application requirements.

Top-down views in precision livestock have been deployed since early work and still have importance for several reasons (Psota *et al*., 2019). Nevertheless, the top-down view approach also presents challenges such as occlusion, shadows, and camera placement. Animals may obstruct each other, leading to partial or full blockage or drastic shape deformation in the images, making accurate individual tracking and recognition challenging. The height at which cameras are mounted can introduce scale variations, impacting algorithm robustness. Furthermore, annotating large-scale livestock datasets with bounding boxes, species labels, behavior categories, and other relevant information can be an extremely resource-intensive task, especially for video data. Access to comprehensive annotated datasets remains limited. The absence of contextual meta information makes it difficult to correlate observations with external factors, such as weather conditions, feeding schedules, or animal health records. This hampers the ability to draw meaningful insights and make informed decisions. In summary, the challenges and prospects in livestock farming computer vision dataset preparation are intrinsically linked to the development and success of precision livestock farming. While challenges exist, they are met with innovative solutions such as transfer learning, image augmentation, and multimodal fusion.

#### *4.3.6. Ethics and protocol for animal experiments*

Animal research has emerged as one of the most sensitive and regulated activities in many countries. Animal welfare is a crucial aspect of using animals in the experiment. Concerns over animal well-being during animal husbandry, trading, and scientific experiments have evolved as serious issues. Fortunately, with significant scientific progress and morale, nowadays, the humane treatment of animals has become a part of the culture in modern society (Silverman, 2015). However, harmonization between scientific goals and animal well-being must exist together. To ensure the safeguarding and rights of animals, an oversight agency that monitors and regulates animal handling has been established and operated in many nations, like the Institutional Animal Care and Use Committee (IACUC) in the United States (U.S.). Getting approval from the IACUC is the first and foremost process for conducting animal-related research in the U.S. The basic requirements for IACUC approval are described in Mohan and Foley (2019).

Although there is a lack of standard forms and protocols used by oversight agencies globally, there are many common grounds for their work (Mohan and Foley, 2019). Generally, the 3Rs (Reduce: reducing the number of animals used for minimum experiment results; Replace: replacing the sentient animals with less sentient or simulated objects as far as possible; and Refine: refining experimental design or techniques to provide appropriate animal husbandry, experimental skills, and pain relief) concept developed by Russell and Burch (1959) are well accepted universally. A team comprising at least scientists, veterinarians, IACUC, and animal care staff is suggested to balance between scientific experiments and animal welfare, minimizing or eliminating non-protocol variables and approval hustles that could adversely affect the scientific validity and repeatability of the study (Weed and Raber, 2005). It is also recommended that researchers present the documentation of imagery curation processes, with adequate metadata, methodologies, and tools used to collect and process data, and animal health and distress conditions, to allow the minimum repetition of animals and enrich the animals' well-being in future research. Nevertheless, collaborative efforts and adherence to ethical frameworks not only drive advancements in precision animal agriculture but also uphold ethical and welfare considerations, ensuring the coexistence of smart technologies and livestock farming.

### 4.4. Way forward

Recent research on PLF is diverging towards multiple dimensions of livestock monitoring and management, with efforts increasingly focused on developing fully automated systems with minimal human intervention. Based on the findings of this systematic study, we have identified some key limitations in prevailing datasets and the ways the research community could step up to mitigate those limitations.

***Limitations***

- There is limited documentation or metadata provided on the contextual information of datasets.
- There is no uniformity in data collection and annotation given similar applications due to a lack of well-accepted protocols and standards for livestock farming.
- Some public datasets do not fully comply with the FAIR guiding principles (Wilkinson et al., 2016).
- Some datasets are hosted in the personal cloud drives of researchers, which could be inaccessible anytime. Moreover, no backup storage of datasets in non-permanent sites can lead to damage or loss of data accidentally.

- There is inadequate data available for quantitative analysis and early detection of animal health status.
- The lack of diverse and heterogeneous datasets may restrict the robustness of associated models.

***Way forward***

- A collective effort from diverse stakeholders (academia, industry, and funding agencies) could establish standards and protocols for livestock data collection and preparation.
- A leading role from accredited organizations is inevitable in building a large-scale benchmarking database for PLF.
- Multi-modal algorithms could be useful in precisely detecting complex animal behaviors and subtle health issues.
- Multi-view camera and imaging techniques could alleviate occlusion issues allowing long-term tracking of animals.
- Advanced 3D imagers with high resolution and ranging accuracy can potentially increase the robustness of body measurement and weight estimation tasks.
- Maintenance and preservation of open-source datasets are necessary to ensure reusability in the future.
- Semi-supervised or self-supervised modeling can help alleviate data annotation challenges (Jaihuni et al., 2023).
- The synthetic data via generative modeling or image augmentation techniques (Li and Tang, 2020; Y. Lu et al., 2022) could be supplemental to real-world imagery.

## 5. Conclusion

Datasets are the cornerstone of progress in PLF, enabling the development and success of data-driven solutions in animal agriculture. This study systematically surveys the landscape of CV datasets for livestock farming. The proliferation of CV technologies in animal agriculture has yielded remarkable advancements with numerous datasets published in the monitoring, management, and welfare assessment of livestock. A total of 58 public datasets were identified, encompassing various livestock species, including 27 cattle datasets, 17 for swine, 5 for poultry, and 9 for other animals. The public datasets were dominated by the application of animal detection and tracking, followed by posture and behavior detection, live body measurement, and health and welfare monitoring applications. More than two-thirds of the datasets were captured by RGB cameras, with the remaining primarily by RGB-D cameras. The challenges such as illumination variations, occlusion, and dataset documentation underscore the complexities of working with livestock data. Nonetheless, new opportunities lie ahead for developing advanced AI-powered CV systems. There is a continued need for dedicated efforts to establish more publicly available, high-quality datasets for specific or emerging applications, the development of real-time monitoring systems, and the intensified exploration of AI techniques in PLF. The availability of large-scale high-quality datasets promises to inspire community efforts to produce transformative outcomes for sustainable and efficient livestock farming with improved animal welfare.

---

[i] https://github.com/Anil-Bhujel/Public-Computer-Vision-Dataset-A-Systematic-Survey